%% file: main.tex
\documentclass[11pt]{article}
\usepackage[preprint]{acl}
\usepackage{times}
\usepackage{latexsym}
\usepackage[T1]{fontenc}
\usepackage[utf8]{inputenc}
\usepackage{microtype}
\usepackage{graphicx}
\graphicspath{{figs/}}
\usepackage{subcaption}
\usepackage{amsmath,amssymb}
\usepackage{booktabs}
\usepackage{array}
\usepackage{algorithm}
\usepackage{algpseudocode}

\newcommand{\ours}{\mbox{SharedSAE}}

\title{\ours{}: One Feature Dictionary Across Language Models}

\author{%
  Daniil Ognev$^{1}$ \quad
  Célian Vasson$^{2}$\quad
  Lijie Hu$^{1}$ \quad
  Kentaro Inui$^{1,3,4}$ \quad
  Benjamin Heinzerling$^{3,4}$ \\[4pt]
  $^{1}$MBZUAI \qquad
  $^{2}$Sorbonne Université \qquad
  $^{3}$Tohoku University \qquad
  $^{4}$RIKEN AIP%
}

\begin{document}
\maketitle
\input{body}

\bibliography{custom}

\clearpage
\input{appendix_modern}

\end{document}

%% file: body.tex
\begin{abstract}
Sparse autoencoders (SAEs) are widely used to interpret language model activations, but SAE training and latent labeling are typically repeated for every model.
Here, we show that a single shared SAE can replace a collection of dedicated per-model SAEs.
Our method, \ours{}, combines a shared dictionary with model-specific encoder--decoder pairs.
Unlike the closest prior method, which discards activation magnitudes and requires all models at inference, \ours{} instead normalizes only selection scores, preserving magnitudes, and uses model dropout for single-model inference.
We train \ours{} on four 1B-scale base language models spanning distinct families and tokenizers.
Despite sharing its latents across models, \ours{} retains 96.6\% of dedicated SAEs' mean explained variance; its latent activations exhibit cross-model correlations 1.8 times as high as separate SAEs aligned post hoc, and its latent descriptions transfer across models.
After the dictionary is frozen, new models can be efficiently adapted to it, achieving near-dedicated-SAE reconstruction quality while reusing the shared latent descriptions.
\end{abstract}

\section{Introduction}
\label{sec:intro}

Motivated by superposition \citep{elhage2022toy}, sparse autoencoders (SAEs) use dictionary learning to decompose language-model activations into sparse latents \citep{bricken2023monosemanticity,cunningham2023sparse,gao2024scaling}.
These latents can be labelled \citep{bills2023language} and used to interpret, monitor, steer, or trace a model's computations \citep{templeton2024scaling,marks2025sparse,deng2026qwenscope}.
Yet, current practice trains an SAE and describes its latents separately for each target model.
For instance, GemmaScope trains many layer- and model-specific SAE suites \citep{lieberum2024gemmascope}, whose dictionaries expose millions of latents requiring automated description \citep{paulo2024automatically}.
A new target model would require another SAE, another description set, and analyses built around another latent space.
We show that a single shared SAE can avoid this duplication and offer a plausible alternative to dedicated per-model SAEs.

A shared SAE must reconstruct each model from the learned shared sparse space despite differences in hidden width, architecture, tokenization, and activation scale.
Its latents must support interpretation and downstream use \citep{makelov2024towards,karvonen2025saebench}, and they must stay fixed with their descriptions when new models are attached. At inference, a shared SAE must only require the model of interest to obtain sparse encodings and reconstructions.  
Existing approaches meet only subsets of these requirements: post-hoc matching retains separate dictionaries, which can differ even on the same data \citep{paulo2025different}, and recovers only partial correspondences \citep{lan2024universal,wang2024towards}.
Earlier shared-dictionary methods instead learn one dictionary jointly across layers or models \citep{lindsey2024crosscoders,thasarathan2025universal,nasirisarvi2025sparc}, but depend on all participating models for inference-time selection: reconstruction degrades when only one model selects latents.
They also cannot attach a new model while keeping the dictionary and its descriptions fixed.

\input{fig_teaser}

Our method, \ours{}, meets these requirements by obtaining one shared dictionary via tokenizer, latent choice, and activation magnitude alignment. Per-model encoder-decoder pairs map each model's hidden width into and
out of the shared dictionary, and tokenizer alignment supplies matching activation
windows across architectures. During joint training, selection scores are
normalized so that TopK picks the same latents regardless of activation scale,
while the selected magnitudes stay untouched, and a cross-stream rescale lets each
model's code reconstruct the others. Model dropout trains every model to pick the
shared latents on its own, which is what makes solo inference possible.
As a result, \ours{} reconstructs model activations more accurately than the earlier shared-dictionary methods.
After joint training, frozen-dictionary adaptation attaches a new model by training only its encoder, decoder, and biases, leaving the shared latents and their descriptions unchanged. Figure~\ref{fig:teaser} summarizes the design against the alternatives

In summary, our contributions are:
\begin{itemize}
    \item \parbox[t]{\linewidth}{We introduce \ours{}, which learns one sparse dictionary across heterogeneous language models, preserves activation magnitudes, and supports single-model inference (Section~\ref{sec:method}).}
    \item \parbox[t]{\linewidth}{We show across four different 1B-scale base models that the shared dictionary retains near-dedicated reconstruction quality, yields stronger cross-model activation alignment than post-hoc matching and prior methods, and supports transferable latent descriptions (Section~\ref{sec:align}).}
    \item \parbox[t]{\linewidth}{We develop frozen-dictionary adaptation, which attaches held-out, post-trained, and architecturally different models without changing the shared latents or their existing descriptions (Section~\ref{sec:adapt}).}
\end{itemize}

\section{Related Work}
\label{sec:related}

\paragraph{Shared dictionaries and model diffing.}
Crosscoders train one sparse dictionary over several activation streams and are the standard instrument of model diffing, comparing a model against its own finetune \citep{lindsey2024crosscoders,minder2025sparsity,mishrasharma2025insights}; \citet{modeldiffing2026} extended diffing across families to find differences.
Shared dictionaries for several models were first built in the vision and multimodal setting: USAE, where one randomly chosen model encodes each batch and all models decode it \citep{thasarathan2025universal}, and SPARC, the closest method to ours, with per-model encoders and decoders, one global top-$k$ over the raw sum of per-stream codes computed on unit-normalized inputs (token magnitudes are discarded at the input and nothing balances the vote), and a cross-reconstruction loss \citep{nasirisarvi2025sparc}.
Neither adapts a new model onto a frozen dictionary, and only USAE supports single-model deployment; both train as baselines under our data and budget (Appendix~\ref{app:baselines}).

\paragraph{Aligning representations after training.}
Whether independently trained networks learn similar representations is measured by metrics like similarity indices \citep{raghu2017svcca,morcos2018pwcca,kornblith2019similarity}, unit matching \citep{li2016convergent,dravid2023rosetta}, stitching maps \citep{lenc2015stitching,bansal2021stitching}, and shared anchors \citep{moschella2023relative}.
Beyond analysis, \emph{post-hoc matching} pairs separately trained SAEs' latents by activation correlation \citep{lan2024universal,wang2024towards}, which needs a shared tokenizer and pairs a minority of latents.
\emph{Concept-atlas alignment} retrofits a new model onto an already-labelled space, either by representational alignment \citep{puri2025atlas} or by fitting a linear decoder onto a panel-trained library \citep{wu2026atlases}; these approaches reuse labels but neither train the space jointly across families nor audit whether it is shared.
Concurrent work attaches per-model linear adapter pairs to a frozen dense activation interface \citep{kim2026adapter}; it shares our labelling-economics motivation but trains no sparse dictionary and audits no sharedness.
We use post-hoc matching throughout the sharedness analysis and both attachment mechanisms in the adaptation analysis (Sections~\ref{sec:align} and~\ref{sec:adapt}).

\paragraph{Hypotheses about shared representations.}
The Platonic Representation Hypothesis suggests that models trained on similar data may learn related representational geometries \citep{huh2024platonic}, although the extent of this convergence is contested \citep{koepke2026plato}.
The Linear Representation Hypothesis proposes that many concepts correspond to linear directions \citep{park2023linear}, consistent with findings from embeddings, steering, probing, and monitoring \citep{mikolov2013linguistic,nanda2023emergent,gurnee2024language}.
Together, these hypotheses offer one possible explanation for why model-specific directions could be organized around shared sparse latents despite different activation spaces.
We treat this explanation as background rather than a premise: the experiments establish only the narrower claim that a shared dictionary can be learned for the evaluated model roster.

\section{\ours{}}
\label{sec:method}
\begin{figure*}[t]
\centering
\includegraphics[width=\textwidth]{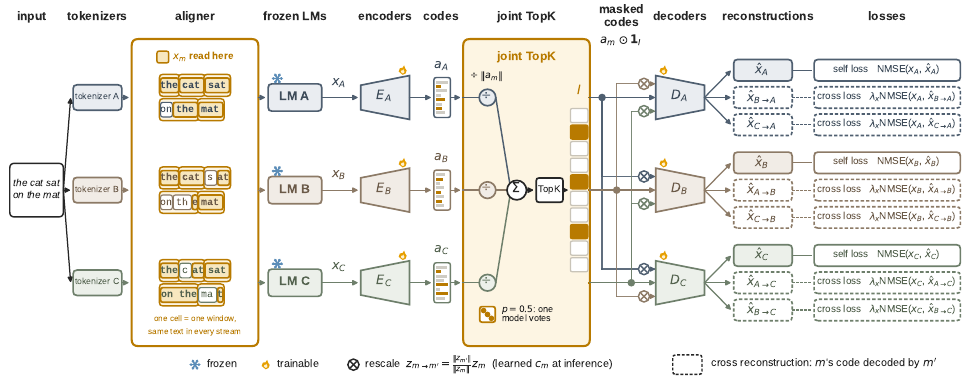}
\caption{SharedSAE. The aligner cuts each model's token stream into windows that
decode to the same text and reads $x_m$ at the last token of each window
(highlighted). Per-model encoders and decoders surround one shared dictionary:
the codes $a_m$ are normalized and summed, one TopK gives the index set $I$ for
all models, and with probability $p = 0.5$ a single random model picks $I$ alone
(dice), which trains solo inference. Each decoder reconstructs its own model
(self loss) and every other model from that model's rescaled code.}
\label{fig:arch}
\end{figure*}

\ours{} is a TopK sparse autoencoder with one dictionary of $L$ latents and a
separate encoder-decoder pair per model (Figure~\ref{fig:arch}). Each text is
tokenized by every model, and an aligner cuts the token streams into windows that
decode to the same text, so that one training tuple holds every model's activation
for the same span. Each model's frozen residual stream then passes through its own
encoder into a dense code over the shared dictionary. The models pick which latents
fire together: their codes are normalized and summed, and one TopK over the sum
gives a single index set for all of them, so every model keeps its own magnitudes
but agrees on the active features. Each decoder reconstructs its own model from the
masked code, and also from every other mode's code after a
magnitude rescale, which ties each latent index to the same feature in every model.
During training a random single model picks the index set half of the time, so that inference-time use of only one model is in-distribution for the architecture.

The building block is the TopK SAE of \citet{gao2024scaling}: $\hat{x} = W_{\mathrm{dec}}\operatorname{TopK}_k(\operatorname{ReLU}(W_{\mathrm{enc}}(x-b)+b_{\mathrm{lat}}))+b$, trained on $\operatorname{NMSE}(x,\hat{x}) = \lVert x-\hat{x}\rVert^2/\lVert x\rVert^2$ plus the auxk dead-latent term $\tfrac{1}{32}\mathcal{L}_{\mathrm{aux}}$, decoder columns renormalized each step so a latent's magnitude lives in its code; dedicated SAEs of this form are our per-model references.

\paragraph{Aligned data.}
Each model reads the same text through its own tokenizer, so the $N$ token sequences differ, where $N$ is the number of models that participate in the training. The aligner walks them in parallel, growing the window whose decoded text is shortest until all $N$ windows decode to the same normalized text - the two-model algorithm of \citet{modeldiffing2026} extended to $N$ (Appendix~\ref{app:align}). Each closed window gives one training tuple $x = (x_1, \dots, x_N)$, where $x_m$ is the residual stream at model $m$'s hook layer.

\paragraph{\ours{}.}
Each model owns an encoder into a shared dictionary of $L$ latents and a decoder out of it:
\begin{equation}
\begin{aligned}
z_m &= W^{m}_{\mathrm{enc}}(x_m - b_m) + b^{m}_{\mathrm{lat}},\\
a_m &= \operatorname{ReLU}(z_m) \in \mathbb{R}^{L}.
\end{aligned}
\label{eq:enc}
\end{equation}
With probability $1-p$ the latents are chosen by the joint TopK (jTopK) of all models' normalized votes, and with probability $p$ a random single model's TopK selects them:
\begin{equation}
I =
\begin{cases}
\operatorname{TopK}_k\!\left(\textstyle\sum_{m=1}^{N} \frac{a_m}{\lVert a_m\rVert + \epsilon}\right), \\
\hfill \text{w.p.\ } 1-p,\\[4pt]
\operatorname{TopK}_k(a_u),\quad u \sim \mathrm{Unif}\{1,\dots,N\}, \\
\hfill \text{w.p.\ } p,
\end{cases}
\label{eq:gate}
\end{equation}
with $p = 0.5$ during training; the second case is \emph{vote-source dropout}. Either way one selection is shared by all models, a constant of the forward pass: gradients flow through the selected entries of every code, none through the selection. At inference one model votes alone - the \emph{solo mode}. $I$ masks each model's code for self-reconstruction, $\hat{x}_m = W^{m}_{\mathrm{dec}}(a_m \odot \mathbf{1}_{I}) + b_m$; the latents are shared, so column $i$ of every $W^{m}_{\mathrm{dec}}$ is model $m$'s direction for shared latent $i$. Cross-reconstruction from model $m$ to $m'$ also rescales the encoding:
\begin{equation}
\begin{aligned}
z_{m \to m'} &= \frac{\lVert z_{m'} \rVert}{\lVert z_m \rVert}\, z_m,\\
\hat{x}_{m \to m'} &= W^{m'}_{\mathrm{dec}}\!\bigl(\operatorname{ReLU}(z_{m \to m'}) \odot \mathbf{1}_{I}\bigr) + b_{m'}.
\end{aligned}
\label{eq:cross}
\end{equation}
The loss adds the self- and cross-reconstruction errors:
\begin{equation}
\begin{aligned}
\mathcal{L} ={}& \frac{1}{N}\sum_{m}\operatorname{NMSE}(x_m,\hat{x}_m) + \tfrac{1}{32}\,\mathcal{L}_{\mathrm{aux}}\\
&+ \frac{\lambda_x}{N(N-1)}\sum_{m \neq m'}\operatorname{NMSE}(x_{m'},\hat{x}_{m\to m'}),
\end{aligned}
\label{eq:loss}
\end{equation}
with $\lambda_x = 1.0$ unless stated.

\paragraph{Architecture Components Rationale.}
We briefly describe the rationale behind the architecture design choices that we introduced that were not found in previous literature. The vote norm in Eq.~\eqref{eq:gate} keeps the model with the largest activation scale from dominating latent selection. The cross term and its rescale allow the latents selected for one model, at their magnitudes, to reconstruct every other model's representation. The rescale reads the target's norm, so a learned per-model scale $c_m$ replaces it at inference. Vote-source dropout makes the architecture inference-friendly, so latents can be obtained faithfully from one model's activations rather than all that trained. Lastly, adaptation: a new model gets a fresh encoder and decoder trained against the latents the original models select, voting alone half the time, with the dictionary frozen since training (Section~\ref{sec:adapt}). Removing any component breaks exactly its target, and only the vote-norm ablation leaves the training loss unchanged (Table~\ref{tab:ablate}).

\section{Testing Sharedness of the Trained Concept Space}
\label{sec:align}

We analyze the learned Shared Linear Concept Space against three criteria: whether the encoders choose the same latents for a token, assign correlated magnitudes, and associate the latents with the same meaning. Baselines are dedicated TopK SAEs matched post hoc by best correlation - an upper bound on any one-to-one assignment - and SPARC and USAE. Every shared system is measured under the joint TopK, where all models vote, and the solo TopK, where one runs alone, like during inference. Some of the metrics we measure may come directly from the loss and alignment pressure components \citep{mishrasharma2025insights}. On every alignment metric, our architecture scores higher than previously proposed variants that sought cross-model alignment. The semantic alignment of the latents, which is the final goal of this interpretability research, was never directly optimized.

\subsection{Common SAE metrics: reconstruction and spliced behavior}
\label{sec:cost}

Sharing costs a few percent of the converged ceilings, and one model's code decodes into every other's space at most of the target's variance, beating the strongest affine stitch between raw streams on all twelve pairs (Tables~\ref{tab:main},~\ref{tab:reconmatrix}). Under the solo TopK \ours{} scores \emph{higher} than under the joint one on three of four streams - alone, each model picks what suits it specifically - and spliced back into the model it recovers nearly all of the mean-ablation cross-entropy gap: within 0.006 mean CE-recovered of the converged dedicated splice (0.957 vs 0.963, Table~\ref{tab:main}), and exceeding the budget-matched 8-epoch dedicated SAEs on three of four streams (Table~\ref{tab:functional}; both metrics are defined in Appendices~\ref{app:def:fve} and~\ref{app:def:splice}).

\subsection{Do the models choose the same latents?}
\label{sec:crit1}
\input{fig_balance}

\input{fig_codemass}

First, we measure how many models each latent serves with $N_{\mathrm{eff}} = \bigl(\sum_m n_m\bigr)^2 / \sum_m n_m^2$, where $n_m$ is the number of tokens on which model $m$'s solo vote selects the latent: $N_{\mathrm{eff}}=4$ when all four models use it equally often, $1$ when one uses it alone. Second, following crosscoder diffing \citep{lindsey2024crosscoders}, we ask the same question of code mass. Writing $\mu_m(i)$ for the share of model $m$'s total code mass spent on latent $i$, which removes the scale difference between streams, each pair of models splits the latent as
\begin{equation}
r_{mm'}(i) = \frac{\mu_m(i)}{\mu_m(i) + \mu_{m'}(i)},
\label{eq:codeshare}
\end{equation}
and we report $\max_{m<m'} \lvert r_{mm'}(i) - \tfrac{1}{2} \rvert$, the largest departure from an even split over the six pairs: 0 means every pair is balanced, and 0.5 means some pair is entirely one-sided. Both readings are defined in Appendix~\ref{app:def:usage}. On both, most \ours{} latents are used by all four models near-equally, while the baselines sit at the far end, where at least one pair carries nearly all the mass - the one-sided use familiar from diffing (Figures~\ref{fig:balance} and~\ref{fig:codemass}).

\input{fig_token_choice}

\input{fig_top50}

Per token, we measure whether each model's encoder selects the same latents for aligned tokens (Figure~\ref{fig:tokenchoice}). We then report how much reconstruction the ``consensus latents'' provide (Table~\ref{tab:consensus}; Appendix~\ref{app:def:choice}).

Per latent, we obtain the top-50 tokens that activate it using each model's own encoder and report the four-way intersection of the four sets (Appendix~\ref{app:def:choice}). SPARC's median collapses to zero through a usage-breadth effect rather than disagreement (Appendix~\ref{app:sparcsolo}), and USAE overlaps only on its frequent core (Figure~\ref{fig:top50}).


\subsection{Do the activation magnitudes correlate?}
\label{sec:crit2}

\input{fig_inversion}

\input{tab_modern_main}

\input{tab_modern_reconmatrix}

Choosing the same latents is not enough for concept alignment. Given a token, encodings from different models can select the same latents but assign unrelated magnitudes to them. For each latent we therefore collect its activations on aligned tokens in all four models, compute the Pearson correlation for the six model pairs, and report the median per latent (Appendix~\ref{app:def:corr}). Figure~\ref{fig:inversion} shows that SPARC attains the higher magnitude correlation when latents are chosen by the joint TopK, where a shared selection makes the models co-fire by construction and part of what is measured is the selection rather than the representation \citep{mishrasharma2025insights,usama2026convergence}. When each model chooses its latents alone, the ordering inverts and \ours{} leads every other architecture: its models select largely the same latents (Section~\ref{sec:crit1}), and the correlated activations indicate that those latents carry the same linear concept across models rather than corresponding only by construction. The post-hoc control bounds what matching alone can achieve, since separately trained SAEs disagree on most latents even across seeds \citep{paulo2025different,lan2024universal}. SPARC's mass of non-correlating latents is examined in Appendix~\ref{app:sparcsolo}: Gemma's solo vote uses roughly a quarter of the dictionary while its other streams use most of it, and because the joint selection is confined to that narrow shared core, the disagreement outside it appears only when the models vote alone.

\vspace{-0.5em}
\subsection{Do the shared latents mean the same things?}
\label{sec:crit3}

\input{semanticity}

\input{functional}

\paragraph{A targeted search finds shared latents.}
Top-activation browsing favors frequent, lexical features, so we also search in the
opposite direction: starting from 13 concept families that the interpretability
literature reports \citep{templeton2024scaling,bricken2023monosemanticity}, we probe
each with eight matched text pairs and admit a latent only if its probe contrast is
positive in \emph{all four} models, ranked by the weakest, and its top held-out
activations independently confirm the concept. Most candidates fail this bar: when a
winner's held-out activations track a topic correlate rather than the concept itself,
it is discarded, and only the survivors are reported. The survivors go beyond the
lexical layer: one evaluative latent responds to the same concept in both probe
languages, and the space holds both poles of a valence axis, in one dictionary shared
by four model families (Figure~\ref{tab:hunt}). Five of the six draw their strongest
activations from dozens of distinct passages per model; the sixth concentrates in the
few documents of its domain, a data-availability caveat we detail with each model's
strongest contexts in Appendix~\ref{app:huntex}.

\input{fig_hunt}

\vspace{-0.3em}
\section{Adaptation of Newly Released Models}
\label{sec:adapt}

We now test adaptation after freezing \ours{}: the original models' joint TopK supplies a reference latent selection for each token, and only the new model's encoder, decoder, and biases are trained. We apply the procedure of Section~\ref{sec:method} unchanged in three kinds of setting: models held out from our roster, a post-trained model, and architectures the seed never saw.
\vspace{-0.3em}
\paragraph{Leave-one-out folds.}
We hold out each roster model in turn, train a three-model seed, and adapt the fourth into it. We compare the performance to what the held-out model performs like when \ours{} is trained with all models present. Deployed solo, every fold is within a few percent of its dedicated ceiling (Table~\ref{tab:adapt}).

\input{tab_modern_adapt}

\paragraph{Qwen3.5: adaptation to a hybrid architecture.}
Qwen3.5-2B-Base is a stronger test \citep{qwen35card}, a hybrid in which three of every four blocks use linear attention. The trained \ours{} never saw this architecture. We report two adaptation runs, with and without vote-source dropout, under which the new model chooses the latents alone half the time. Both match a five-model co-trained \ours{} under the joint TopK, but only the dropout run lets the model select latents from its own code alone, which solo deployment requires.

\paragraph{A post-trained model joins the base-model space.}
The same recipe attaches Llama-3.2-1B-Instruct, a post-trained checkpoint whose base
sibling is already in the seed, to the frozen dictionary: deployed solo it reaches
1.006 of its own dedicated SAE (Table~\ref{tab:adapt}), so the labels learned on base
models carry over to this instruct model without loss of reconstruction. A common transformer from an
unrelated lab passes the same test: Hunyuan-1.8B \citep{hunyuancard}, released after the
dictionary was trained, adapts to 1.010 of its own dedicated SAE and reproduces the
probe-found latents' contexts (Appendix~\ref{app:huntex}).

\input{tab_modern_adaptbattery}

\paragraph{Adapted members share the space like co-trained ones.}
We check whether adaptation only matches the reconstruction of the co-trained run, or whether the adapted model joins the existing shared latents and holds up on the sharedness criteria of Section~\ref{sec:align} (Table~\ref{tab:adaptbattery}). Voting solo, the folds match the co-trained dictionary on magnitude correlation and on top-50 overlap, and they land higher on the per-token intersection. Qwen3.5 ties its five-model reference on correlation and clears it on choice and on FVE share. We compare against the two attach mechanisms of Section~\ref{sec:related}, instantiated on our own frozen dictionary: a linear alignment into a seed stream \citep{puri2025atlas} and a decoder-only attach \citep{wu2026atlases}. Our adapted encoder-decoder pair leads both on every criterion we measure, so the adapted model joins the dictionary rather than reading out of it. The same holds latent by latent, and semantically: on the six probe-found latents of Figure~\ref{tab:hunt}, all three adapted models fire on the frozen latents' contexts as the seed streams do (value correlations of 0.64--0.99 with every seed stream, top-50 position overlaps up to 44 of 50), so each adapted model inherits the meaning each latent had when it was labelled (Appendix~\ref{app:huntex}).

\section{Conclusion}
\label{sec:conclusion}

We tested whether a Shared Linear Concept Space that different LLMs map into is obtainable, and on our roster every criterion fixed in advance is met: the dictionary reconstructs each stream within a few percent of its converged dedicated ceiling and, spliced, is within 0.006 mean CE-recovered of the converged dedicated splice, exceeding budget-matched dedicated SAEs on three of four streams (Section~\ref{sec:align}). The space is shared on all three criteria: the models choose the same latents, assign correlated magnitudes, and give them the same meaning. When a new model is adapted to it, it adopts the existing vocabulary better than any other methods and can be used alone (Section~\ref{sec:adapt}).

\section*{Limitations}
\label{sec:limits}

\ours{} is sensitive to how it is trained, and we vary one ingredient at a time rather
than search the space. Every number we report moves with the amount of aligned data and
with the training budget and the layer we read. The roster is hooked at
roughly 85\% depth, and the same recipe at mid depth gives different cross-model
reconstruction (Table~\ref{tab:mid} in Appendix~\ref{app:protocol:tables}), so depth is a free parameter we fixed once
rather than tuned. Since layers at different depths serve different purposes in the LLMs, the resulting dictionary and sharedness statistics might differ in unexpected ways. The roster is bounded the same way: four base checkpoints around 1B
parameters on English FineWeb-Edu, which leaves alignability at larger scale and in
other languages untested; post-training is now probed once, with Llama-3.2-1B-Instruct
adapting onto the frozen base-model dictionary at 1.006 of its own dedicated SAE
(Table~\ref{tab:adapt}), but a single instruct checkpoint does not necessarily establish the general case.
Run-to-run variance, by contrast, is measured and small: three converged dictionary
seeds agree to within 0.003 FVE per stream, joint and solo. Beyond these, the architecture has more
moving parts than we can ablate. The vote norm, the dropout probability, the
cross-reconstruction weight, the rescale, the dictionary width and $k$ all interact, and
Table~\ref{tab:ablate} varies them one at a time around a single settled configuration.
We do not claim that configuration is optimal, only that removing any one component degrades the property it was introduced for.

Metrics for SAEs are mostly proxies. The property of interest is whether a label written once holds in every model; the measurable quantities are reconstruction, behavior under
a splice, agreement of selections and magnitudes, and whether a generated description
survives being scored on another model. Each is a proxy, and a system could score well on
all of them while the labels still failed to transfer, which is why we report several
rather than one. Two of the three sharedness criteria are also optimized by training;
Section~\ref{sec:align} states that objection with its answers, and
Section~\ref{sec:adapt} is the test it cannot reach. Cross-reconstruction FVEs use the
oracle per-token rescale of Section~\ref{sec:method}, which reads the target model's
pre-activation norm at evaluation time; the learned per-stream scale $c_m$ is the
inference-safe variant and concedes a few points of share, concentrated on the
extreme-scale stream (Table~\ref{tab:ablate}), while self-reconstruction and the
solo-TopK splice involve no rescaling. Pooled FVE and corpus-level CE average over
tokens, so behavior on rare tokens is under-weighted, and the per-latent correlation is
computed on support the selection shares by construction. We therefore corroborate it
with statistics that do not depend on the shared selection: the top-50 overlap against
the matched control (Table~\ref{tab:overlap}), rare latents whose top activations decode
to identical tokens under four different tokenizers (Appendix~\ref{app:examples}), and a
probe with no TopK selection that measures raw encoder responses directly
(Appendix~\ref{app:taxonomy}). A single measure of alignment that depends on no part of
the shared machinery remains future work. Corpus dependence, by contrast, is now
measured: re-evaluating the converged dictionary on a Wikipedia store it never saw leaves
the sharedness criteria essentially unchanged (Appendix~\ref{app:xmodel}).

The dictionary is also not a one-concept-per-latent inventory: we observe feature
splitting, where ``United'' surfaces as two separate latents, so latent counts overstate
concept counts; a fixed shared $k$ can itself entangle features when mis-set
\citep{chanin2025sparse}.

%% file: fig_teaser.tex
\begin{figure*}[t]
\centering
\includegraphics[width=\textwidth]{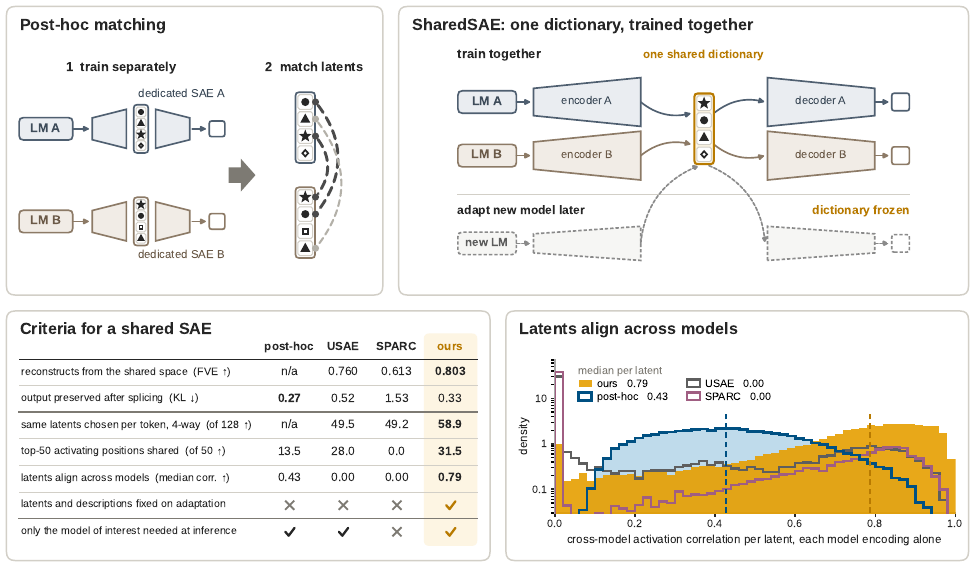}
\caption{\ours{} replaces post-hoc matching with one persistent, adaptable
dictionary. Top left: dedicated SAEs admit only partial post-hoc matching of
their latents. Top right: \ours{} trains per-model encoders and decoders around
one shared dictionary and later attaches a new model with the dictionary frozen.
Bottom left: the criteria a shared SAE must meet, with each method's score.
Bottom right: under solo TopK, \ours{}'s latents correlate across models far
more strongly than post-hoc matched SAEs, USAE, or SPARC.}
\label{fig:teaser}
\end{figure*}

%% file: fig_balance.tex
\begin{figure}[!ht]
\centering
\includegraphics[width=\columnwidth]{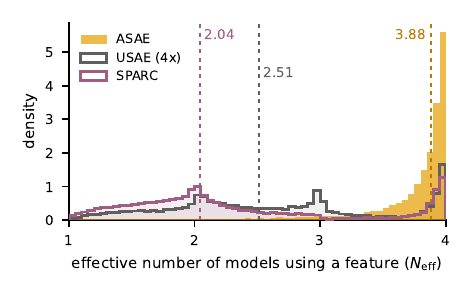}
\caption{How many models use a latent: $N_{\mathrm{eff}}$ - 4 when all four models
select it equally often, 1 for single-model use.}
\label{fig:balance}
\end{figure}

%% file: fig_codemass.tex
\begin{figure}[!ht]
\centering
\includegraphics[width=\columnwidth]{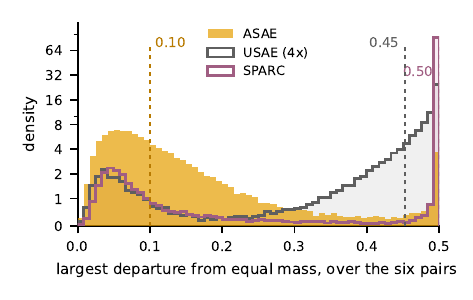}
\caption{The share of a latent's code mass one model contributes within a pair,
pooled over the six pairs and symmetrized.}
\label{fig:codemass}
\end{figure}

%% file: fig_token_choice.tex
\begin{figure}[!ht]
\centering
\includegraphics[width=\columnwidth]{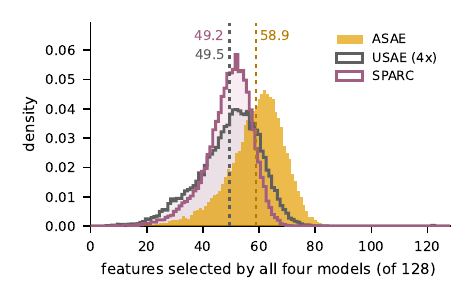}
\caption{Latent selection under the solo TopK, per token: each model's encoder alone
selects its top-128 latents for the same aligned token; the distributions count how
many of those latents all four models select at once. Dotted lines mark the means.}
\label{fig:tokenchoice}
\end{figure}

%% file: fig_top50.tex
\begin{figure}[!ht]
\centering
\includegraphics[width=\columnwidth]{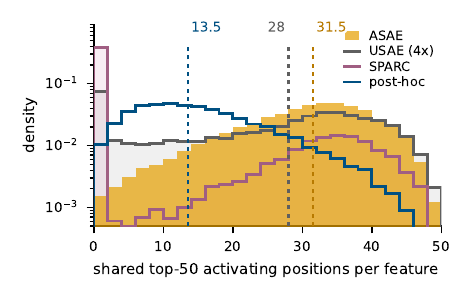}
\caption{Latent selection under the solo TopK, per latent: we obtain the top-50 tokens
that activate each latent and report the four-model intersection of these four sets.}
\label{fig:top50}
\end{figure}

%% file: fig_inversion.tex
\begin{figure}[t]
\centering
\includegraphics[width=\columnwidth]{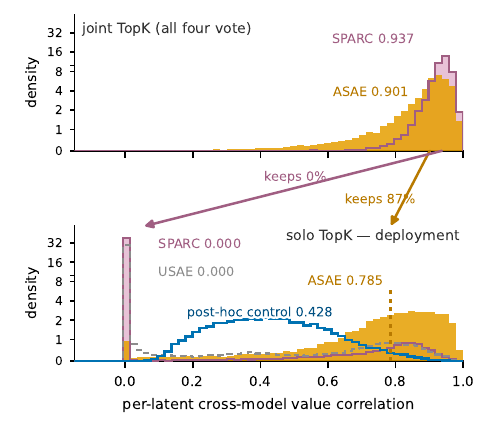}
\caption{Per-latent cross-model value correlation under the joint TopK (top),
where all four models vote for the latents to use, and under the solo TopK (bottom),
where each model chooses alone. The labels are each system's median}
\label{fig:inversion}
\end{figure}

%% file: tab_modern_main.tex
\begin{table*}[t]
\centering
\small
\setlength{\tabcolsep}{5pt}
\begin{tabular}{lcccccc}
\toprule
 & Qwen3 & Llama-3.2 & OLMo-2 & Gemma-3 & Mean & Cross-FVE \\
\midrule
\multicolumn{7}{l}{\emph{Reconstruction (FVE)}} \\
Dedicated (ceiling) & 0.838 & 0.807 & 0.812 & 0.868 & 0.831 & - \\
\ours{}, joint TopK & 0.798 & 0.785 & 0.787 & \textbf{0.825} & 0.799 & \textbf{0.749} \\
\ours{}, solo TopK & \textbf{0.804} & \textbf{0.794} & \textbf{0.795} & 0.821 & \textbf{0.803} & - \\
USAE & 0.760 & 0.743 & 0.750 & 0.787 & 0.760 & 0.640 \\
SPARC, joint TopK & 0.769 & 0.765 & 0.763 & 0.631 & 0.732 & 0.663 \\
SPARC, solo TopK & 0.747 & 0.728 & 0.727 & 0.251 & 0.613 & - \\
Crosscoder & 0.679 & 0.690 & 0.693 & 0.233 & 0.574 & - \\
\midrule
\multicolumn{7}{l}{\emph{CE-recovered under the solo-TopK splice $\uparrow$}} \\
Dedicated (ceiling) & 0.974 & 0.937 & 0.971 & 0.970 & 0.963 & \\
\ours{}, solo TopK & \textbf{0.958} & \textbf{0.933} & \textbf{0.970} & \textbf{0.968} & \textbf{0.957} & \\
USAE & 0.925 & 0.885 & 0.941 & 0.950 & 0.925 & \\
SPARC, solo TopK & 0.923 & 0.899 & 0.933 & 0.325 & 0.770 & \\
Crosscoder & 0.942 & 0.919 & 0.937 & 0.729 & 0.882 & \\
\midrule
\multicolumn{7}{l}{\emph{$\mathrm{KL}(\text{clean}\,\|\,\text{patched})$ $\downarrow$}} \\
Dedicated (ceiling) & 0.186 & 0.399 & 0.262 & 0.246 & 0.274 & \\
\ours{}, solo TopK & \textbf{0.309} & \textbf{0.438} & \textbf{0.279} & \textbf{0.293} & \textbf{0.330} & \\
USAE & 0.503 & 0.712 & 0.471 & 0.375 & 0.515 & \\
SPARC, solo TopK & 0.501 & 0.613 & 0.522 & 4.501 & 1.534 & \\
Crosscoder & 0.477 & 0.570 & 0.569 & 1.860 & 0.869 & \\
\bottomrule
\end{tabular}
\caption{Held-out reconstruction and spliced behavior, every system at its own
converged budget, under the joint TopK and the solo TopK. For CE-recovered and KL, the
reconstruction replaces the residual stream and next-token behavior is scored.
SPARC's codes are unit-norm, so its printed reconstruction uses the oracle per-token
rescale into each model's space; with the deployable one-scalar-per-model median-norm
rescale its means drop to 0.643 (joint) and 0.490 (solo). Blanks are due to
architectural differences: USAE and the crosscoder have no joint/solo split, the crosscoder has no
cross-reconstruction.
The dedicated rows are the per-model ceiling, a best-case reference. Bold marks the best shared system per column within each block excluding the dedicated reference.}
\label{tab:main}
\end{table*}

%% file: tab_modern_reconmatrix.tex
\begin{table}[t]
\centering\scriptsize
\setlength{\tabcolsep}{3.5pt}
\begin{tabular}{lcccc}
\toprule
 & \multicolumn{4}{c}{decoded into model $m'$} \\
\cmidrule(lr){2-5}
code from $m$ & Qwen3 & Llama-3.2 & OLMo-2 & Gemma-3 \\
\midrule
Qwen3 & \textbf{0.798} & 0.738 & 0.738 & 0.782 \\
Llama-3.2 & 0.746 & \textbf{0.785} & 0.738 & 0.786 \\
OLMo-2 & 0.741 & 0.733 & \textbf{0.787} & 0.780 \\
Gemma-3 & 0.739 & 0.733 & 0.736 & \textbf{0.825} \\
\midrule
Dedicated & 0.815 & 0.783 & 0.795 & 0.825 \\
\bottomrule
\end{tabular}
\caption{Self- and cross-reconstruction FVE of \ours{} under the joint TopK.}
\label{tab:reconmatrix}
\end{table}

%% file: semanticity.tex
\paragraph{Shared latents retain the interpretability of dedicated latents.}
A shared dictionary is only useful if its latents remain interpretable across models. We compare two labeling procedures. Under \emph{shared-vote labeling}, examples are ranked by the shared activation vote and a single description is assigned to each shared latent. Under \emph{per-stream labeling}, examples are ranked separately by each model's latent activations, producing one description per latent and model. We evaluate every description against the corresponding target-stream activations using the FADE framework \citep{puri-etal-2025-fade}, which measures clarity, responsiveness, purity, and faithfulness (Appendix~\ref{app:fade}). Because FADE scores are only meaningful when the description correctly identifies the latent, we first analyze label--stream pairs satisfying a clarity threshold of $0.9$; retained coverage should therefore be interpreted as part of the labeling protocol, not only as a property of the learned dictionary.

Within this retained set, \ours{} shared-vote labels achieve FADE scores comparable to dedicated per-model SAEs (Table~\ref{tab:fade}, top). Their retained rate is also in the same range as the dedicated baselines under the same automatic labeling protocol. Per-stream labels retain more pairs, but this setting gives each shared latent four model-specific labeling attempts, whereas shared-vote labeling assigns one description to the latent. Thus, on a clear and qualitative label, the shared dictionary does not show a substantial loss of interpretability relative to independently trained SAE dictionaries.

\input{tab_fade}

\paragraph{Shared latents admit transferable semantic descriptions.}
We next evaluate whether descriptions that identify a latent in one model transfer to the other models. We consider two source-validation strategies (Table~\ref{tab:fade}, middle). For \emph{\ours{} shared vote}, a latent is retained if the shared label reaches clarity $\geq0.9$ in at least one stream, and this same label is evaluated on all target streams. For \emph{\ours{} clearest per-stream}, we first select, for each latent, the per-stream label with the highest source-stream clarity and retain the latent only when this value exceeds $0.9$; the selected label is then evaluated across all streams.
Source-validated descriptions transfer across architectures. Shared-vote labels retain substantial semantic quality when evaluated on other streams, while the clearest per-stream labels obtain slightly higher transfer scores despite being selected using model-specific examples. The transfer scores decrease relative to source-stream validation but remain substantial, suggesting that these descriptions capture properties of the shared latent rather than only stream-specific activation artifacts.

To evaluate semantic consistency beyond the retained-label subset, we remove the target-stream clarity filter and select, for each latent, the per-stream label with the highest source-stream clarity. We then correlate the resulting per-latent FADE scores across the six model pairs (Table~\ref{tab:fade}, bottom). Clarity, responsiveness, and purity show strong cross-model correlations, indicating that latents receiving strong semantic evaluations in one architecture tend to receive similarly strong evaluations in others.
Together, these results support a qualified semantic conclusion: \ours{} shared latents that receive clear qualitative labels obtain FADE scores comparable to dedicated SAE latents, and labels validated in one stream transfer substantially to the other streams. This supports reusable semantic descriptions for shared \ours{} latents, while leaving full-coverage automatic labeling as a separate limitation.
\vspace{-0.5em}
\paragraph{Functional alignment is partial.} Semantic transfer does not imply functional equivalence. Two models may associate the same shared latent with similar contexts while using that information differently during prediction. Table~\ref{tab:fade} reflects this distinction: clarity, responsiveness, and purity are strongly correlated across streams, whereas faithfulness is lower and less consistent.

We therefore treat functional alignment as a weaker and more conservative claim than semantic transfer. The quantitative results indicate that shared latents preserve some cross-model functional structure, but they do not establish latent-level causal equivalence. Appendix~\ref{fig:steer_examples} provides qualitative steering examples in which the same shared latent produces related behavioral shifts across models.

%% file: tab_fade.tex
\begin{table}[t]
\centering
\scriptsize
\setlength{\tabcolsep}{2.5pt}
\resizebox{\columnwidth}{!}{%
\begin{tabular}{lccccc}
\toprule
 & Clarity & Resp. & Purity & Faith. & Retained \\
\midrule
\multicolumn{6}{l}{\emph{Within-stream clear labels (clarity $\geq 0.9$)}} \\
\ours{} shared vote & 0.979 & 0.642 & 0.463 & 0.281 & 10.6\% \\
\ours{} per-stream & 0.976 & 0.629 & 0.438 & 0.217 & 21.7\% \\
SAE Qwen3 & 0.975 & 0.584 & 0.416 & 0.333 & 7.4\% \\
SAE Llama-3.2 & 0.962 & 0.558 & 0.431 & 0.403 & 9.4\% \\
SAE OLMo-2 & 0.954 & 0.583 & 0.373 & 0.320 & 4.1\% \\
SAE Gemma-3 & 0.970 & 0.625 & 0.451 & 0.414 & 10.2\% \\
\midrule
\multicolumn{6}{l}{\emph{Cross-stream transfer}} \\
\ours{} shared vote & 0.850 & 0.569 & 0.401 & 0.207 & 17.2\% \\
\ours{} clearest stream & 0.894 & 0.597 & 0.422 & 0.205 & 30.0\% \\
\midrule
\multicolumn{6}{l}{\emph{Mean cross-stream correlation}} \\
Shared-vote labels & 0.929 & 0.919 & 0.919 & 0.371 & -- \\
Clearest-stream labels & 0.911 & 0.904 & 0.893 & 0.459 & -- \\
\bottomrule
\end{tabular}%
}
\caption{FADE evaluation of \ours{} and dedicated SAE latents.
Top: within-stream evaluation of label--stream pairs satisfying clarity $\geq0.9$. \emph{Retained} reports the fraction of evaluated label--stream pairs satisfying the clarity threshold.
Middle: cross-stream transfer of source-validated labels. Shared-vote labels are retained if they reach clarity $\geq0.9$ in at least one stream and are then evaluated on all streams. Clearest per-stream selects, for each latent, the per-stream label with highest source-stream clarity and retains it when this value exceeds $0.9$.
Bottom: no clarity filtering is applied. For each latent, the per-stream label with highest source-stream clarity is selected and FADE scores are correlated across the six model pairs.}
\label{tab:fade}
\end{table}

%% file: functional.tex
Beyond the faithfulness scores of Table~\ref{tab:fade}, we test functional relatedness directly: we compute input and output scores from \citep{arad-etal-2025-saes} for labelled shared latents and correlate them across model pairs. The resulting correlations are positive but moderate, with mean correlations of 0.52 for input scores and 0.48 for output scores across the six model pairs. These values provide evidence that shared latents tend to play related functional roles across models. Thus, our functional evidence is best interpreted as partial alignment. \ours{} does not directly optimize latent-level causal effects; its objective encourages reconstruction through shared latents. The observed score correlations therefore suggest that the learned latents are not merely semantically aligned labels, but also preserve some functional structure across streams. At the same time, the moderate correlations leave substantial room for model-specific downstream use.


Further probes bound the alignment from both sides. On the positive side, the same
latent causally promotes overlapping output vocabularies far above a mismatched-latent
null, one blind label fits all four models for 88 of 90 mined latents, a targeted
probe search confirms shared high-level latents up to a cross-lingual valence axis
(Figure~\ref{tab:hunt}), and steering the same latent shifts every model toward the
same concept at matched doses (Appendix~\ref{app:xmodel}). On the negative side, the
causal-effect agreements sit far below perfect agreement, token-level detector
features never steer, and the concept--fluency trade-off is similar but not identical
across models. Shared latents give a common handle for intervention, while the
computation each model performs with them stays partly model-specific.

%% file: fig_hunt.tex
\begin{figure*}[t]
\centering
\includegraphics[width=\textwidth]{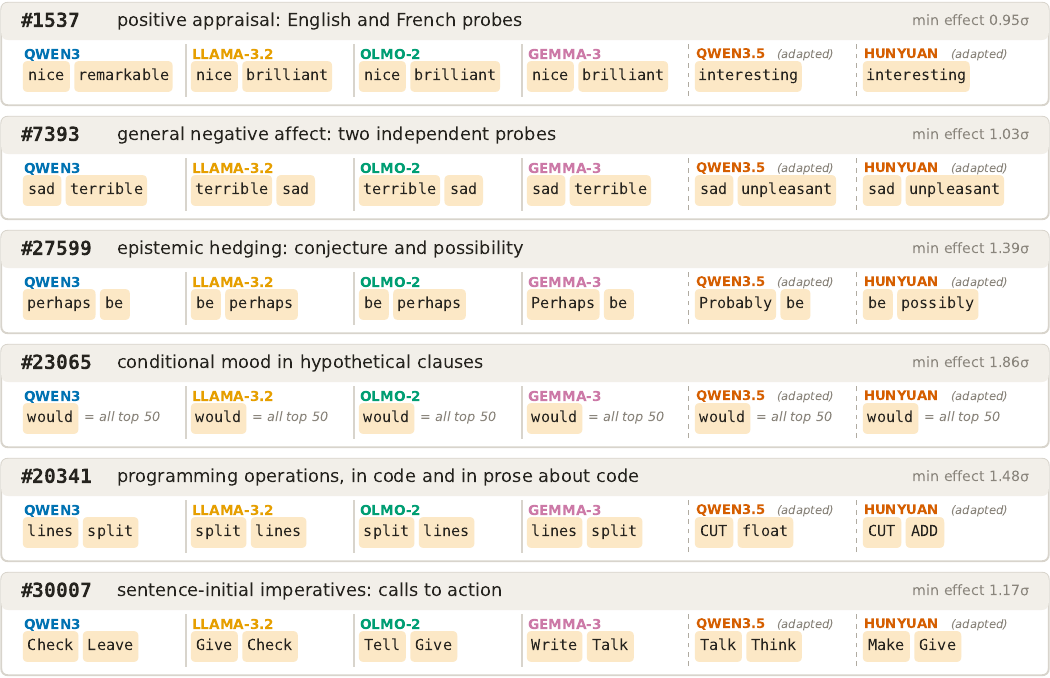}
\caption{Most-activating tokens for selected single-concept latents, across the four
models \ours{} was trained on and two models adapted to the frozen dictionary
afterwards. \emph{min effect} is the probe gap in the weakest model, in
standard-deviation units. Full contexts are in Appendix~\ref{app:huntex}.}
\label{tab:hunt}
\end{figure*}

%% file: tab_modern_adapt.tex
\begin{table}[t]
\centering
\footnotesize
\setlength{\tabcolsep}{2.5pt}
\begin{tabular}{lccccc}
\toprule
 & \multicolumn{3}{c}{Seed TopK} & \multicolumn{2}{c}{Solo} \\
\cmidrule(lr){2-4}\cmidrule(lr){5-6}
Held-out model & FVE & /co-tr. & /ded. & FVE & /ded. \\
\midrule
\multicolumn{6}{l}{\emph{Leave-one-out (three-model seed)}} \\
Qwen3 & 0.769 & 0.997 & 0.944 & 0.793 & 0.972 \\
Llama-3.2 & 0.753 & 0.990 & 0.962 & 0.769 & 0.982 \\
OLMo-2 & 0.751 & 0.984 & 0.945 & 0.776 & 0.977 \\
Gemma-3 & 0.787 & 0.979 & 0.954 & 0.812 & 0.984 \\
\midrule
\multicolumn{6}{l}{\emph{Qwen3.5-2B (hybrid; four-model seed)}} \\
~~seed-only (protocol) & 0.751 & 0.990 & 0.937 & 0.349 & 0.436 \\ 
~~vote-source dropout & 0.751 & 0.990 & 0.937 & 0.779 & 0.972 \\ 
\midrule
\multicolumn{6}{l}{\emph{Llama-3.2-1B-Instruct (post-trained; four-model seed)}} \\
~~vote-source dropout & 0.743 & -- & 0.973 & 0.768 & 1.006 \\
\midrule
\multicolumn{6}{l}{\emph{Hunyuan-1.8B (cross-lab transformer; four-model seed)}} \\
~~vote-source dropout & 0.791 & -- & 1.004 & 0.795 & 1.010 \\
\bottomrule
\end{tabular}
\caption{Adapting a held-out model to a frozen \ours{}: only the new model's encoder,
decoder, and biases train, so the seed models and their latent descriptions stay untouched.}
\label{tab:adapt}
\end{table}

%% file: tab_modern_adaptbattery.tex
\begin{table}[H]
\centering
\scriptsize
\setlength{\tabcolsep}{2.6pt}
\begin{tabular}{lcccc}
\toprule
 & corr. & top-50 & choice & FVE/ded. \\
\midrule
Co-trained \ours{} (4 models) & 0.785 & 31.5 & 59.0 & 0.999 \\
Adapted, leave-one-out mean & 0.785 & 31.5 & 66.0 & 0.979 \\
\midrule
\multicolumn{5}{l}{\emph{Held-out Gemma fold, attach routes}} \\
\quad adapted encoder-decoder (ours) & 0.785 & 32.0 & 60.3 & 0.984 \\
\quad linear alignment (Atlas-style) & 0.722 & 29.5 & 57.5 & 0.692 \\
\quad decoder-only attach & - & - & - & 0.795 \\
\midrule
\multicolumn{5}{l}{\emph{Qwen3.5 (five streams)}} \\
\quad co-trained, five-model & 0.784 & 32.0 & 58.1 & 0.960 \\
\quad adapted encoder-decoder (ours) & 0.784 & 31.0 & 59.7 & 0.972 \\
\quad linear alignment (Atlas-style) & 0.757 & 30.0 & 51.3 & 0.670 \\
\quad decoder-only attach & - & - & - & 0.881 \\
\midrule
\multicolumn{5}{l}{\emph{Llama-3.2-1B-Instruct (five streams)}} \\
\quad adapted encoder-decoder (ours) & 0.786 & 31.0 & 51.8 & 1.006 \\
\bottomrule
\end{tabular}
\caption{The Section~\ref{sec:align} criteria under adaptation, with the adapted model voting solo:
median per-latent value correlation and top-50 overlap against the seed models, mean
per-token intersection of the solo selections (of 128; four-way above, five-way in the
Qwen3.5 and Instruct blocks), and solo FVE as a share of the dedicated SAE.}
\label{tab:adaptbattery}
\end{table}

%% file: appendix_modern.tex
\appendix

\input{fig_curves}
\input{tab_modern_ablate}
\input{tab_modern_functional}
\input{tab_modern_consensus}
\input{tab_modern_overlap}
\input{tab_modern_features}

\section{Experimental setup}
\label{sec:setup}

One roster, one aligned corpus, one recipe, metrics fixed in advance; the few exhibits kept at an earlier configuration disclose it in place.

\paragraph{Models.}
Four separately released base checkpoints from four families with four tokenizers: Qwen3-1.7B-Base \citep{qwen3report}, Llama-3.2-1B \citep{grattafiori2024llama}, OLMo-2-0425-1B \citep{olmo2report}, gemma-3-1b-pt \citep{gemma3report}, hooked at roughly 85\% depth (mid depth in Table~\ref{tab:mid} in Appendix~\ref{app:protocol:tables}). The roster is sink-free there, with no massive-activation outliers \citep{sun2024massive,sun2026spike} to distort a variance-weighted metric, but median token norms differ by three orders of magnitude - the regime the vote norm is designed for.

\paragraph{Data.}
Activations come from FineWeb-Edu \citep{penedo2024fineweb}, aligned as in Section~\ref{sec:method}: 4M positions, of which 3.9M train; shard~9 is evaluation-only for every system here, and a separate 200k store extends the held-out window to 300k positions for per-latent statistics.

\paragraph{Training.}
Every system trains at $L=32{,}768$, TopK $k=128$ on the same data; \ours{} and the dedicated references share one trainer (auxk, AdamW \citep{loshchilov2019decoupled}, unit-normed decoder columns; details in Appendix~\ref{app:metrics}), USAE and SPARC their own reference recipes. At the matched 8-epoch budget every validation curve is still rising (Figure~\ref{fig:curves}), so every system trains until validation FVE levels off under one early-stopping criterion capped at 64 epochs: Table~\ref{tab:main} compares converged against converged (Appendix~\ref{app:baselines}).

\paragraph{Baselines and metrics.}
The dedicated ceiling is one TopK SAE per model on the same trainer with RMS-calibrated inputs; these four converged ceilings are Table~\ref{tab:main}'s reference rows and the denominator of every share we quote. Reconstruction is the fraction of variance explained (FVE), centered and pooled, on held-out shard~9; function is tested by splicing reconstructions in at the hook layer under the solo TopK, since at inference only one model runs, reporting CE-recovered and $\mathrm{KL}(\text{clean}\,\|\,\text{patched})$ \citep{karvonen2025saebench}. SPARC's reconstructions are unit-norm, so its raw-space FVE uses the oracle rescale. Formulas, counts, and the baseline arms behind each exhibit are in Appendices~\ref{app:metrics} and~\ref{app:baselines}.

\section{Artifacts, licenses, and data}
\label{app:licenses}

\paragraph{Models.}
All five checkpoints are publicly released base models, used through the Hugging Face
Hub and only for extracting residual-stream activations for research on
interpretability, which is within the stated terms of each. Qwen3-1.7B-Base
\citep{qwen3report} and Qwen3.5-2B-Base \citep{qwen35card} are released under Apache
2.0, as is OLMo-2-0425-1B \citep{olmo2report}. Llama-3.2-1B \citep{grattafiori2024llama}
is released under the Llama 3.2 Community License, and gemma-3-1b-pt
\citep{gemma3report} under the Gemma Terms of Use. Both of the latter are custom
licenses that permit research use and carry use-based restrictions; we comply with them
and redistribute no model weights.

\paragraph{Data.}
Activations are taken from FineWeb-Edu \citep{penedo2024fineweb}, which is released
under the Open Data Commons Attribution License (ODC-By 1.0) and is filtered from
Common Crawl web snapshots. The corpus is English, and the paper's claims are therefore
limited to English; behavior in other languages is untested. We use a small fraction of
it: 4M token positions that survive the $N$-way alignment of
Appendix~\ref{app:align}, of which 3.9M are used for training and the remainder for
evaluation, with shard~9 held out from every system in this paper. We do not
redistribute the corpus, and we apply no filtering beyond the alignment procedure, so
the documents carry whatever the FineWeb-Edu educational-quality filter admits. The
text is public web data that was not collected by us and contains no information about
the authors or any identified individual that we introduce; we did not screen it for
personally identifying content beyond the source filtering, which is a limitation
inherited from the corpus.

\paragraph{Compute.}
Every run in this paper uses a single NVIDIA RTX PRO 6000 Blackwell Server Edition GPU
on a shared cluster; no run is distributed across GPUs. The shared dictionary holds
$L = 32{,}768$ entries with one encoder-decoder pair per model, at residual widths 2048
for Qwen3, Llama-3.2 and OLMo-2 and 1152 for Gemma-3, which is 478.4M trainable
parameters for the four-model \ours{}, the same budget as four dedicated SAEs of matched
width. The target models are frozen throughout and contribute no trainable parameters.
Individual runs are short: the four leave-one-out adaptation folds take 45 to 49 minutes
each and the five-stream Qwen3.5 arms 67 minutes, while the converged four-model \ours{} run
takes about four hours (job log: epochs at $\sim$11 minutes under the 64-epoch cap with
early stopping). The experiments reported here, including every baseline, every
convergence run, and the exploratory runs that did not reach the paper, account for
roughly 260 GPU-hours in the cluster's job records.

\paragraph{Code.}
SPARC is the authors' released implementation \citep{nasirisarvi2025sparc}, run on our
store converted to its expected input format. USAE is our own reimplementation of the
published recipe \citep{thasarathan2025universal}, since no code release was available.
The crosscoder and the two attach routes are our implementations of the published
descriptions, each verified against the numbers reported in the original work before
being used to produce new ones (Appendix~\ref{app:baselines}).

\paragraph{Software versions.}
\ours{}, the dedicated references, the crosscoder and both attach routes are trained
and scored under Python 3.12.13 with PyTorch 2.12.0.dev20260408 built against CUDA
12.8, Transformers 5.9.0, Tokenizers 0.22.2, Safetensors 0.7.0, Datasets 4.8.5, NumPy
2.4.6 and SciPy 1.17.1; figures are drawn with Matplotlib 3.10.9. The two published
baselines run in their own environments so that each matches its reference recipe: SPARC
under Python 3.11.15 with PyTorch 2.12.1 built against CUDA 13.0, from the authors'
repository at commit \texttt{7a9cfc1}, modified only to read our activation store and
to change the epoch count; USAE under Python 3.12.13 with the same PyTorch 2.12.1 build,
from our reimplementation, since no code release was available. Model weights are
obtained through the Hugging Face Hub at the revisions current in July 2026.

\section{Metric definitions}
\label{app:metrics}

This section defines every metric the paper reports. Each definition states the data it
is computed on, the quantity formed per position or per latent, and the number we
report. Section~\ref{app:protocol} then gives the window, population, and settings used
for each individual exhibit.

\paragraph{Notation.}
Let $i = 1, \dots, N$ index the aligned positions of the evaluation window, $m = 1,
\dots, M$ the models ($M = 4$ unless stated), and $j = 1, \dots, L$ the entries of the
shared dictionary ($L = 32{,}768$). Model $m$ reads position $i$ as the residual stream
vector $x_m(i) \in \mathbb{R}^{d_m}$, and its encoder turns that into a non-negative
code before selection,
\begin{equation}
a_m(i,j) = \bigl[\,W^m_{\mathrm{enc}}(x_m(i) - b_m) + b^m_{\mathrm{lat}}\,\bigr]_j^+ .
\label{eq:appenc}
\end{equation}
A selection rule returns a set $I(i) \subset \{1,\dots,L\}$ with $|I(i)| = k$, and the
sparse code keeps the selected entries only,
\begin{equation}
z_m(i,j) = a_m(i,j)\,\mathbb{1}\bigl[\,j \in I(i)\,\bigr].
\label{eq:appcode}
\end{equation}
Two selection rules are used throughout. The \emph{joint TopK} takes the $k$ largest
entries of the summed normalized votes of all models, $I^{\mathrm{joint}}(i)$, and
is the same set for every model. The \emph{solo TopK} lets one model select on its own,
$I^{\mathrm{solo}}_m(i) = \operatorname{TopK}_k a_m(i,\cdot)$, which differs from model
to model and is the deployment condition. Reconstructions are written $\hat{x}_m(i)$.
A latent is \emph{alive} for a model when that model selects it at least $\tau$ times
in the window; thresholds are stated per exhibit in Section~\ref{app:protocol}.

\subsection{Reconstruction}
\label{app:def:fve}

Training minimizes a per-position, uncentered error, $\mathrm{NMSE} =
\operatorname{mean}_i \lVert x_m(i) - \hat{x}_m(i) \rVert^2 / \lVert x_m(i) \rVert^2$.
Evaluation uses the pooled, centered fraction of variance explained,
\begin{equation}
\mathrm{FVE}_m = 1 - \frac{\sum_i \lVert x_m(i) - \hat{x}_m(i) \rVert^2}
                          {\sum_i \lVert x_m(i) - \bar{x}_m \rVert^2},
\label{eq:appfve}
\end{equation}
where $\bar{x}_m = \frac{1}{N}\sum_i x_m(i)$ is the mean activation over the window.
\emph{Cross-FVE} is Eq.~\eqref{eq:appfve} with $\hat{x}_{m'}(i)$ produced from model
$m$'s code, rescaled as in Section~\ref{sec:method} and passed through model $m'$'s
decoder; we report the mean over all $M(M-1)$ ordered pairs. \emph{Share of ceiling}
divides a system's $\mathrm{FVE}_m$ by the dedicated SAE's $\mathrm{FVE}_m$ on the same
window.

\subsection{Behavior when the reconstruction is spliced in}
\label{app:def:splice}

We replace the hook layer's activations with the reconstruction and run the model
forward. Writing $\mathrm{CE}^{\mathrm{clean}}_m$ for the untouched next-token
cross-entropy, $\mathrm{CE}^{\mathrm{patch}}_m$ for the value after the replacement,
and $\mathrm{CE}^{\mathrm{abl}}_m$ for the value when the activations are replaced by
the dataset mean vector,
\begin{equation}
\mathrm{CE\text{-}rec}_m =
\frac{\mathrm{CE}^{\mathrm{abl}}_m - \mathrm{CE}^{\mathrm{patch}}_m}
     {\mathrm{CE}^{\mathrm{abl}}_m - \mathrm{CE}^{\mathrm{clean}}_m},
\label{eq:appce}
\end{equation}
so 1 means the reconstruction leaves behavior unchanged and 0 means it is worth no more
than the mean vector. We also report the divergence between the two next-token
distributions, $\mathrm{KL}_m = \operatorname{mean}_i \mathrm{KL}\bigl(p^{\mathrm{clean}}_m(\cdot \mid i)
\,\big\|\, p^{\mathrm{patch}}_m(\cdot \mid i)\bigr)$, for which lower is better.

\subsection{How many models use a latent}
\label{app:def:usage}

Let $n_m(j) = \lvert \{\, i : j \in I^{\mathrm{solo}}_m(i) \,\} \rvert$ be the number of
positions at which model $m$'s own vote selects latent $j$. The effective number of
models using that latent is the inverse Simpson index of those four counts,
\begin{equation}
N_{\mathrm{eff}}(j) = \frac{\bigl(\sum_m n_m(j)\bigr)^2}{\sum_m n_m(j)^2}
\; \in \; [1, M],
\label{eq:appneff}
\end{equation}
which equals $M$ when all models select the latent equally often and 1 when a single
model selects it alone.

The same question can be asked of code mass rather than selection counts. Let
$\mu_m(j) = \sum_i z_m(i,j) \big/ \sum_{j'} \sum_i z_m(i,j')$ be the share of model
$m$'s total code mass that lands on latent $j$, which removes the scale difference
between streams. Each pair of models then splits the latent as
\begin{equation}
r_{mm'}(j) = \frac{\mu_m(j)}{\mu_m(j) + \mu_{m'}(j)},
\label{eq:appshare}
\end{equation}
and we report the largest departure from an even split over the six pairs,
$\max_{m<m'} \lvert r_{mm'}(j) - \tfrac{1}{2} \rvert$: 0 means every pair is balanced,
and 0.5 means some pair is entirely one-sided.

\subsection{Whether the models select the same latents}
\label{app:def:choice}

\paragraph{Per position.}
Each model selects its own $k$ latents at position $i$. We count how many are selected
by all four at once,
\begin{equation}
C(i) = \Bigl\lvert \bigcap_{m=1}^{M} I^{\mathrm{solo}}_m(i) \Bigr\rvert
\; \in \; \{0, \dots, k\},
\label{eq:appagree}
\end{equation}
and report the mean of $C(i)$ over positions. The latents in that intersection are the
\emph{consensus latents} of the position; Table~\ref{tab:consensus} reports the FVE
obtained when only they are kept and the remaining $k - C(i)$ entries are zeroed.

\paragraph{Per latent.}
For latent $j$ and model $m$, let $S_m(j)$ be the set of 50 positions at which
$a_m(i,j)$ is largest, so $\lvert S_m(j) \rvert = 50$. The four models give four such
sets, and we report the size of their intersection,
\begin{equation}
O(j) = \Bigl\lvert \bigcap_{m=1}^{M} S_m(j) \Bigr\rvert \; \in \; \{0, \dots, 50\},
\label{eq:apptop50}
\end{equation}
taking the median of $O(j)$ over the alive latents. Under independent selection the
expected value is $50\,(50/N)^{M-1}$, which is below $10^{-3}$ at our window sizes.
Pairwise overlap replaces the intersection over four models with one over a single pair.
\emph{Token purity} is the largest share of any one token type among the 50 positions in
$S_m(j)$.

\subsection{Whether the activation magnitudes agree}
\label{app:def:corr}

For latent $j$ and models $m \neq m'$ we correlate the two sparse codes across the
window, keeping positions where either model is zero,
\begin{equation}
\rho_{mm'}(j) = \frac{\operatorname{cov}_i\bigl(z_m(i,j),\, z_{m'}(i,j)\bigr)}
                     {\sigma_m(j)\,\sigma_{m'}(j)} .
\label{eq:appcorr}
\end{equation}
Each latent is summarized by the median of $\rho_{mm'}(j)$ over the six model pairs in
which it is alive on both sides, and each system by the median of that quantity over
latents. Zeros are kept because a latent that one model fires and another does not is
a genuine disagreement, and dropping those positions would hide it.

The \emph{post-hoc matched control} applies Eq.~\eqref{eq:appcorr} to separately trained
dedicated SAEs, whose latents are distinct. For each latent of model $m$'s dictionary we take
the latent of model $m'$'s dictionary with which it correlates best, with no one-to-one
constraint and with both entries required to fire at least 10 times. This is an upper
bound on what any one-to-one assignment between two independently trained dictionaries
could achieve.

\subsection{Selection diagnostics}
\label{app:def:selection}

A model's \emph{vote share} at a position is $\lVert v_m \rVert / \sum_{m'} \lVert
v_{m'} \rVert$, where $v_m$ is its contribution to the summed joint vote, averaged over
positions. \emph{Solo match} is the fraction of the joint selection a model recovers on
its own, $\lvert I^{\mathrm{solo}}_m(i) \cap I^{\mathrm{joint}}(i) \rvert / k$, averaged
over positions. The \emph{solo-vocabulary Jaccard} compares, between two models, the
sets of latents each selects on at least $0.1\%$ of positions.

\emph{Sink share} is the fraction of positions whose token RMS exceeds eight times the
median; it is $0.000\%$ on this roster.

\input{fig_convergence}

\section{Measurement protocol}
\label{app:protocol}

This section states, for each exhibit, the window it is measured on, the population of
latents or positions it covers, and any setting that is not the default. The metrics
themselves are defined in Section~\ref{app:metrics}.

\subsection{Windows and populations}
\label{app:protocol:windows}

Self-FVE in Table~\ref{tab:main} is centered and pooled over the first 20,000 positions
of held-out shard~9, which no system in this paper trains on. Cross-FVE is the mean over
all twelve ordered model pairs. Per-latent statistics use a longer window of 300k
aligned positions, formed from held-out shard~9 plus a separate 200k evaluation store.
The splice measurements of Eq.~\eqref{eq:appce} run on 160 FineWeb-Edu sequences of 256
tokens; clean cross-entropy is 2.379, 2.427, 2.736 and 2.519 for Qwen3, Llama, OLMo-2
and Gemma, and 2.428 for Qwen3.5.

For latent usage and code mass (Figures~\ref{fig:balance} and~\ref{fig:codemass}) we
count solo-TopK selections on the 300k window and keep latents selected at least 50
times in total: 26,838 latents for \ours{}, 31,983 for SPARC and 24,490 for USAE. The
value correlation of Eq.~\eqref{eq:appcorr} uses the same window with a threshold of 10
selections per pair.

\subsection{Convergence of the references}
\label{app:protocol:convergence}

Early stopping fired on Llama at epoch 53, its validation peak unbeaten for eight
consecutive epochs. Qwen3, OLMo-2 and Gemma reach the 64-epoch cap, gaining at most
0.005 FVE over the final doubling of the budget and showing no sustained decline. A
second Qwen3 seed moves its ceiling by 0.001. A 32-epoch run with a decaying learning
rate is 0.005 above the constant-rate ceiling on this window, which is the gain
available from the schedule that every system here shares, so the ceilings stay
comparable. \ours{}, under the same criterion, stops at epoch 62 with its peak
at epoch 54. These statistics are read from the trainer's validation series over the
full held-out shard, while the printed ceilings are rescored on the 20,000-position
window of Table~\ref{tab:main}; the two windows differ by 0.5 to 1.0 points of FVE.
Figure~\ref{fig:convergence} shows the full curves. The dedicated references train on
RMS-calibrated inputs; retraining them on raw inputs at the earlier protocol moved each
ceiling by at most 0.002 FVE, so the calibration does not inflate them.
Tables~\ref{tab:ablate} and~\ref{tab:mid} divide by that protocol's own 10-epoch
references.

SPARC's reconstructions are unit-norm. Its raw-space FVE in Table~\ref{tab:main} scales
each one by the position's true norm, which uses information from the target. Scaling
instead by the stream's median norm, which does not, gives mean self-FVE 0.643 under the
joint TopK, 0.490 under the solo TopK, and cross-FVE 0.569.

\subsection{Cross-model activation-correlation protocol}
\label{app:protocol:teaser}

The population is every latent that fires in at least two models, 27,514 of 32,768,
measured on 100k held-out positions. The median on this window is 0.89, against the 0.90
that Section~\ref{sec:crit2} reports on the 300k window. The post-hoc control pairs each
dedicated SAE's latent with its best-correlated partner in another model's SAE, and its
median of 0.44 is taken over all six pairs. Under the same protocol the converged
dedicated SAEs give a pooled best-match median of 0.450, with per-pair medians from
0.399 to 0.506. Matching two converged Qwen3 SAEs trained from different seeds gives
0.770 best-match and 0.871 mutual-best, which is the best the post-hoc route achieves
within a single model, on one seed pair and one stream. Figure~\ref{fig:yardstick} draws
the three distributions together.

\subsection{Controls}
\label{app:protocol:controls}

Every pipeline below reproduces its published numbers before producing new ones. The
affine-stitch, identity and weight-space items are measured on the 8-epoch dictionary.

\paragraph{Floor without a dictionary.}
Matching raw residual dimensions across models by best correlation on the 300k window
gives a pooled median $\lvert r \rvert$ of 0.14, or 0.13 under the signed convention,
and is unchanged when the massive-activation dimensions are excluded. This is the zero
line beneath Figure~\ref{fig:yardstick}.

\paragraph{Linear map between raw streams.}
A ridge map fit on 1.7M training positions with tuned regularization reaches mean
cross-FVE 0.610, against 0.726 for the shared code, and is below it on all twelve
ordered pairs by 0.085 to 0.179. The gap is widest where the two streams are least
linearly related.

\paragraph{Permutation nulls.}
One hundred frequency-matched draws, permuting latents within per-model usage
deciles, place chance at about 0 value correlation, 0 of 50 top-50 overlap, about 3 of
128 per-position selections, and 0.012 modal-token agreement. Every headline number
exceeds all one hundred draws, a one-sided empirical $p = 0.0099$. The two usage-balance
statistics have high chance levels by construction, 3.65 and 3.43 of 4, which the
observed 3.88 and 3.90 clear.

\paragraph{Token groups.}
Four-way selection agreement stays between 52 and 60 of 128 when positions are grouped
by token frequency, by content class, and by position in the sequence. The lowest cell
of any interaction is 49, on a small group. Rare content tokens keep about nine tenths
of the headline value.

\paragraph{Do the shared latents give the best map?}
Dense ridge and sparse matching-pursuit maps between two models' solo code spaces never
beat the direct map between shared latents with a per-latent rescale: total $R^2$ is 0.66 to 0.70 for the
free maps against 0.69 to 0.74 for the direct map, on all six pairs.

\paragraph{Decoding one model's code into another.}
Encoded by one model under its solo TopK and decoded by another model's decoder, with a
fixed per-pair scale estimated on training shards, the spliced code recovers 0.92 to
0.96 of the target's cross-entropy gap, mean 0.94. Using the per-position rescale
instead changes this by at most 0.015, so that rescale does not carry the result. The
target's own code, spliced at the same positions, recovers 0.95 to 0.97. Crossing models
therefore costs about 0.015 of CE-recovered, although the divergence to the clean model
roughly doubles, 0.45 to 0.65 against 0.27 to 0.32.

\paragraph{Agreement in the weights.}
Decoder columns are unit-normed by construction, so the relative-norm signal used in
crosscoder diffing is degenerate here. Decoder \emph{directions} still agree above a
shuffled-pairing floor, SVCCA 0.34 against 0.17 and latent RSA 0.31 on the core of
latents all four models use, against 0.00. This is evidence that does not depend on the
TopK at all, and it is weaker than the selection-based evidence.

\subsection{Figure~\ref{fig:curves}}
\label{app:protocol:curves}

Validation FVE by epoch, at a constant learning rate. Every dedicated run and the shared
\ours{} rise monotonically through epoch 8, which is why every system is trained to
convergence. The epoch 41 to 62 tail in the middle and right panels is the converged
\ours{} run, a separate stream with seed 301, on the same full-shard window. A node requeue
truncated that run's log for epochs 1 to 40; its resume line pins epoch 40 at best mean
self-FVE 0.7882, continuous with the plotted tail.

\subsection{Figure~\ref{fig:inversion}}
\label{app:protocol:inversion}

Histograms are density-normalized so that populations of different size compare fairly.
Under the joint TopK, SPARC's median is 0.937 on 7,194 latents and \ours{}'s is 0.901
on 25,374. Under the solo TopK \ours{}'s median is 0.785 on 27,439 latents, with 86\%
above 0.5, against 35\% for the post-hoc control at median 0.428 on 32,756. \ours{}'s
mass near zero is 925 latents, 3.4\% of the measured population, carrying 0.1\% of all
solo selections. These are rare latents, selected a median of 76 times per 300k
positions against 1,359 for the latents above 0.5, and they are not single-model
latents: their usage spreads over 2.7 models at the median and only 6\% sit below
$N_{\mathrm{eff}} = 1.5$. The model-specific latents are treated separately
in Appendix~\ref{app:taxonomy}. Weighted by use, 96\% of solo selections fall on
latents correlating above 0.5. SPARC's solo median is 0.000 on 31,812 latents and its
distribution is bimodal, with the three Gemma pairs at medians 0.740 to 0.770 and the
other three at 0.000. USAE trains per-model codes natively, so it has no joint or solo
variant; its median is 0.000 on 25,342 latents.

\subsection{Tables of the main comparison}
\label{app:protocol:tables}

\paragraph{Table~\ref{tab:main} and Table~\ref{tab:functional}.}
Windows and clean cross-entropy values are given in
Section~\ref{app:protocol:windows}. Solo TopK means a single model's own vote picks the
latents, which is the deployment mode.

\paragraph{Table~\ref{tab:features}.}
Usage classes are defined at a threshold of $0.1\%$ of positions, on windows of 20k and
100k positions. The correlation column is the median per-latent cross-model code
correlation within the class.

\paragraph{Table~\ref{tab:adapt}.}
The two ratio columns divide by a co-trained \ours{} that saw the held-out model from the
start, and by that model's dedicated SAE. Solo means the new model's own vote picks the
latents. Every adaptation run uses $k = 128$, 8 epochs and held-out shard~9. On one GPU
the four leave-one-out folds took 45 to 49 minutes each and the five-stream Qwen3.5 arms
67 minutes. Qwen3.5-2B-Base is a hybrid model in which three of every four blocks use
linear attention, Gated DeltaNet, and the fourth uses softmax attention; we hook layer
20 of 24 and align it into a five-model store of 4M positions. The first arm keeps the
seed-only TopK throughout training. The second gives the new model's own vote sole
control of the selection with probability $0.5$. The adapted model's median code correlation
to the frozen seed models is 0.897 to 0.911 in the second arm and 0.901 to 0.915 in the
first.

\paragraph{Table~\ref{tab:ablate}.}
Share is the mean self-FVE share of the dedicated ceilings, and the remaining columns are
the median value correlation and the median top-50 overlap across all four models, all at
the earlier protocol stated in the caption. In the vote-input block only the third row
uses the vote norm, and vote shares are rounded to two decimals. With raw votes Gemma
keeps 0.956 of its ceiling while the other three drop to 0.90 to 0.92; with
RMS-calibrated inputs the vote shares still spread by a factor of six, Gemma's at 0.068.

\paragraph{Table~\ref{tab:overlap}.}
The upper block is computed on the 100k held-out shard. For the pairwise column we take
every latent alive at least 10 times in both members of a pair, count how many of its 50
top-activating positions coincide, and report the median over latents, then the median
across the six pairs with the range in parentheses; the per-pair populations run from
19,471 to 19,648 for \ours{}. The all-four column requires the positions to coincide in
all four models at once, on the core of 6,869 latents alive at least 50 times in every
model; 97.0\% of those \ours{} latents overlap at 10 or more of 50. The lower block repeats
the pairwise count on the 300k window under both selection rules, over latents with at
least 50 selections in both members of a pair, and prints each system's population. The
USAE entry is its 8-epoch arm in the upper block and its 32-epoch arm in the lower.
Chance overlap is 0 of 50 at these windows.

Because the populations differ in size, we also recompute the solo medians on
usage-matched cores of equal size, taking each system's 13,065 most-selected measurable
latents, which is the smallest population: \ours{} reaches 33.0 of 50, USAE 28.0,
SPARC 19.0 and the matched dedicated control 13.0. Split by usage decile, \ours{}'s
weakest tenth still exceeds the control's strongest; SPARC's agreement is confined to its
top two deciles and USAE's falls away below its frequent core. Post-hoc matching has no
four-model version, because it has no latents shared by all four models, so those cells
are marked with dashes.

\paragraph{Semantic agreement (Section~\ref{sec:crit3}).}
On held-out shard~9, at the token level, we decode each latent's ten strongest
activating positions in each model's own tokenizer and compare both the most frequent
decoded token and the decoded token set across models, reporting medians over the six
pairs. At the concept level we label positions in ten content categories and measure how
much average precision is retained when each model's five most selective concept
detectors are read out in the other models. The column giving average precision in the
detector's own model states its base selectivity, so retention near 1 is not automatic,
and retention above 1 means the detector is more selective in the receiving model than in
its own. Pooled populations are 600 latents for \ours{} and 591 for the matched
control.

\paragraph{Table~\ref{tab:mid}.}
The mid-depth hooks are layer 14 of Qwen3, 8 of Llama, 8 of OLMo-2 and 13 of Gemma, at
roughly half depth. At the earlier protocol of that table the mid-depth roster reaches a
cross-FVE mean of 0.652 with a maximum of 0.743, against 0.602 for the late-depth roster
at the same protocol.

\paragraph{Table~\ref{tab:examples}.}
Latents are selected on 20 to 99 selections per 100k held-out positions, with their top
15 activating positions coinciding 15 of 15 across all four models. The cells show each
model's top activating tokens, decoded by its own tokenizer.

\input{fig_yardstick}

\section{Token alignment procedure}
\label{app:align}

Different tokenizers split the same text differently: one model may hold ``1989'' as a
single token while another holds ``198'' and ``9''. Our aligner is the greedy
decoded-text window-matching algorithm of \citet{modeldiffing2026} (their Algorithm~1),
extended from two models to $N$ (Algorithm~\ref{alg:align}). It walks all $N$ token sequences of the same text chunk
in parallel. At the current positions it first attempts a one-to-one match of the decoded
tokens; on a mismatch it enters window expansion and repeatedly grows the window of the
stream whose decoded text is currently shortest, until all $N$ windows decode to the same
text. When the windows close, we keep each model's activation at the last token of its
window, where self-attention has summarized the matched text, and every model's token id
at that position, for later decoding. Training and evaluation use only these aligned positions.

We deviate from \citet{modeldiffing2026} in four ways. First, the match is $N$-way: all
four models (five during adaptation) close each window simultaneously rather than
pairwise. Second, decoded windows are compared after normalization (Unicode NFKC,
casefolding, and whitespace collapsing), which absorbs tokenizer conventions that differ
in case, spacing, or Unicode form. Third, special tokens and pure-whitespace tokens are
skipped in every stream before matching, so they neither anchor a match nor block one
from closing. Fourth, windows are capped at 16 tokens per stream; if no match is found
under the cap, the remainder of the chunk is dropped and alignment restarts on the next
chunk, keeping every window matched so far. On FineWeb-Edu the cap is rarely reached: in
the 4M production run, 11,495 of 11,504 chunks aligned fully and 9 partially (99.92\%),
and in the earlier 800k-position run all 2,318 chunks completed with zero divergences. Model forwards run in bf16 and
activations are stored in fp32; Gemma-3 cannot run a float16 forward.

\paragraph{Formal statement.}
Algorithm~\ref{alg:align} states the matcher; the guards appear in the order
the implementation applies them, so a window's text is compared before its
length is checked against the cap. One aligned position is emitted per closed
window in every model, at the window's last token.

\begin{algorithm}[H]
\caption{$N$-way greedy decoded-text window matching, the $N$-model
extension of Algorithm~1 of \citet{modeldiffing2026}. $\mathcal{T}_m$ is
model $m$'s tokenizer; $\mathrm{dec}_m(a,b)$ decodes the token span
$t_m[a],\dots,t_m[b{-}1]$; $\nu$ normalizes decoded text (Unicode NFKC,
casefolding, whitespace collapsing); non-content tokens are special or
pure-whitespace tokens; $W{=}16$ is the per-stream window cap. The lagging
stream is the one whose decoded window text is currently shortest, among
streams that can still grow ($e_m < |t_m|$).}
\label{alg:align}
\algrenewcommand\algorithmicindent{1.0em}
\begin{algorithmic}[1]
\Require text $s$; tokenizers $\mathcal{T}_1,\dots,\mathcal{T}_N$; cap $W$
\Ensure kept positions $I_1,\dots,I_N$, one entry per closed window
\State $t_m \gets \mathcal{T}_m(s)$;\; $p_m \gets 0$;\; $I_m \gets ()$ \;$\forall m$
\While{$p_m < |t_m|$ \textbf{for all} $m$}
  \State advance each $p_m$ past non-content tokens
  \If{some stream is exhausted} \State \Return $(I, \textsc{complete})$ \EndIf
  \If{$\nu(\mathrm{dec}_m(p_m,p_m{+}1))$ all agree}
    \State append $p_m$ to $I_m$;\; $p_m{\gets}p_m{+}1$ \;$\forall m$
  \Else \Comment{window expansion}
    \State $e_m \gets p_m{+}1$ \;$\forall m$
    \Loop
      \State $w_m \gets \nu(\mathrm{dec}_m(p_m,e_m))$ \;$\forall m$
      \If{all $w_m$ equal} \Comment{close}
        \State append $e_m{-}1$ to $I_m$;\; $p_m{\gets}e_m$
        \State \textbf{break}
      \EndIf
      \If{$e_m-p_m > W$ for some $m$} \State \Return $(I, \textsc{partial})$ \EndIf
      \State grow the lagging stream: $e_m \gets e_m{+}1$
      \If{no stream can grow} \State \Return $(I, \textsc{partial})$ \EndIf
    \EndLoop
  \EndIf
\EndWhile
\State \Return $(I, \textsc{complete})$
\end{algorithmic}
\end{algorithm}

\paragraph{Measured behavior.}
A tokenizer-only replay of the production extraction's alignment pass, over
the same chunker stream to the same 4M-position target (11,505 contiguous
chunks), reproduces the run: 11,496 chunks align fully and 9 partially
(99.92\%), at 347.7 aligned positions per chunk. The four tokenizers spend a
similar number of tokens per chunk (content-token means 374/364/369/367 for
Qwen3/Llama/OLMo-2/Gemma), and alignment is almost always trivial. Of the
4.0M closed windows, 93.0\% close on the single-token fast path; per-model
window lengths have median 1 and 95th percentile at most 2 tokens (means
1.075/1.048/1.060/1.054), and the longest window ever closed is 16 tokens,
twice in 4.0M. Aligned positions amount to 91.2\%/93.3\%/91.1\%/91.6\% of
each model's raw token count: the closed windows consume 96.5-98.1\% of raw
tokens, special and pure-whitespace tokens account for another 1.9-3.4\%,
and the chunk remainders dropped behind the 9 failed windows cost 0.05\%.

All nine failures hit the window cap, and they concentrate in two text
classes. Seven are non-Latin scripts that the byte-fallback tokenizers split
inside multi-byte UTF-8 codepoints - five Khmer phrases, one pointed
Hebrew word, one Lao royal name: the partially decoded codepoint renders as
the replacement character U+FFFD, which cannot match any other stream until
the codepoint completes, and a run of such codepoints, each costing two or
three tokens in the byte-level stream, exceeds the cap. The other two are
17-digit DOI suffixes (e.g.\ \texttt{10.1590/S1020-49892007000800003}):
Qwen3 and Gemma hold one digit per token, so no unbroken digit run longer
than the cap can close. The same two classes govern the windows that get
long without failing: of the 168 windows (0.004\%) in which any stream
reaches 8 tokens, 136 are digit runs (ISBNs, Unix timestamps, DOIs), 21 are
diacritic-heavy Latin, IPA, or polytonic Greek, 10 are CJK, and one is a
dimension string of multiplication signs and no-break spaces.

Adaptation adds one requirement: the seed store must not change when a model joins. Each
aligned position therefore stores its character-offset anchor in the source text. A new model
re-tokenizes the same chunks and maps its tokens onto the frozen window boundaries; positions
it cannot cover are masked out for its loss terms only. The seed models'
stored activations stay bit-identical, which matches on the data side the frozen
dictionary of Section~\ref{sec:adapt}.

\section{Baseline adaptation to the LLM setting}
\label{app:baselines}

USAE and SPARC were introduced outside the LLM setting, so both are adapted to our roster. USAE is our reimplementation of the published recipe: per-model encoders and decoders around one shared dictionary of the same width as ours; at each step one model is drawn at random to encode the batch, all models decode it, and the self- and cross-reconstruction errors are backpropagated; this coupling follows the released implementation step for step, and where the publication is silent the optimizer settings follow our trainer's conventions (AdamW with weight decay 0.01, a completing warmup-cosine schedule, auxk in place of the always-on auxiliary term). It trains on the same aligned store as \ours{}, for 8, 32, and 64 epochs. SPARC is the authors' released implementation trained on our store converted to its expected input format, under its reference recipe with the dictionary size and $k$ matched to ours - batch 1024 and $k_{\mathrm{aux}}=256$, our roster-wide conventions, replacing its 256 and 64 - and with the held-out shard excluded from training. Its mean-norm and oracle-rescale readout arms share identical codes, so the per-latent measurements of Section~\ref{sec:align} carry one SPARC entry. Both baselines are scored by the protocols of Appendix~\ref{app:metrics}.

Both baselines are trained to the 64-epoch cap of the ceilings, under recipes verified line by line against their released implementations, changing only the epoch count. The budget trajectories show diminishing returns: USAE's mean self-FVE rises 0.604 $\to$ 0.725 $\to$ 0.760 across its 8-, 32-, and 64-epoch arms, flat over the final epochs, and SPARC's rises 0.702 $\to$ 0.732 under the joint TopK and 0.565 $\to$ 0.613 under the solo from 8 to 64 epochs, with Gemma's solo stream still collapsed (0.136 $\to$ 0.251). Table~\ref{tab:main} carries the 64-epoch arms; the per-latent measurements of Section~\ref{sec:align} use the 8-epoch SPARC and 32-epoch USAE arms.

The two attach routes of Table~\ref{tab:adaptbattery} are implemented as follows, with maps and decoders fit on training shards only and every pipeline verified before scoring (an identity-map run reproduces the seed stream's own published statistics exactly, and the decoding path reproduces a published cross-reconstruction cell). \emph{Linear alignment}: a map from the new model's activations into the Qwen3 stream, in two variants - the concurrent recipe's orthogonal Procrustes on row-normalized pairs, and a stronger ridge fit with tuned regularization; the new model then votes solo through Qwen3's encoder, and reconstruction is the round trip through Qwen3's decoder and a reverse map fit the same way. The table reports the stronger ridge variant, with correlations and overlap excluding the mapping-target stream, whose pair is inflated by construction; the Procrustes variant scores lower throughout. \emph{Decoder-only attach}: a fresh decoder trained on the frozen seed models' joint-TopK codes under the standard trainer conventions, with a learned per-source scalar in place of the oracle rescale because this route has no encoder pre-activation norm; its reported reconstruction is the best single source's.

\section{Sharedness across held-out rosters}
\label{app:loo}

\input{fig_loo_sharedness}
\input{tab_modern_loo}

Each leave-one-out adaptation of Section~\ref{sec:adapt} starts from a three-model
seed trained without the held-out model. If sharedness were a fragile property of the
particular four-model roster, these seeds would show it; instead, every seed is as
internally shared as the full dictionary (Table~\ref{tab:loo},
Figure~\ref{fig:loo}). The pooled correlation medians and top-50 overlaps of all four
seeds sit at the full model's level, and the per-token selection intersections are
higher, as expected with fewer voters. Solo FVE shares trail the converged four-model
row only by the budget gap disclosed in the caption. The measurement pipeline was
gated on a self-check: run on the four-model \ours{} restricted to three streams, it
reproduces every published per-pair median exactly.

\section{SPARC's solo agreement}
\label{app:sparcsolo}

\input{tab_modern_sparcsolo}

Two of SPARC's solo statistics have medians of exactly zero (Table~\ref{tab:overlap},
Figure~\ref{fig:inversion}). A zero could mean that the four models select unrelated
latents once the joint vote is removed. Table~\ref{tab:sparcsolo} tests this
reading.

Per token, SPARC's four solo selections agree: the four-way intersection is 49 of
128, against 59 for \ours{}, and every pair shares 59 to 69 latents. No single
stream dominates the selection: vote shares are near-uniform (0.21 to 0.28), and
each stream's own top-$k$ reproduces 0.59 to 0.71 of the joint top-$k$, close to
the 0.73 to 0.76 of \ours{}.

The streams differ in how many latents they use. Over the scoring window, Gemma's
solo vote selects 7,719 distinct latents and OLMo-2's selects 31,097, a factor of
four, and Gemma's set almost coincides with the jointly selected set (7,599). The
four streams of \ours{} use 24,328 to 24,964 latents each. This asymmetry produces
the zeros. A per-latent statistic is computed over the latents active in both
members of a pair. A pair that contains Gemma is therefore measured on the small,
heavily used core, where SPARC agrees (top-50 overlap 32 to 34 of 50, value
correlation 0.74 to 0.77). A pair of two wide-usage streams is measured over 15,000
to 31,000 latents dominated by a rarely firing tail, where both statistics are
zero. The pooled median is zero because the tail-dominated pairs contain two to
four times as many latents.

SPARC therefore keeps per-token agreement and per-latent agreement on a core of
about 7,600 latents, a quarter of its dictionary, and shows no measurable
per-latent agreement outside that core. The agreement of \ours{} is measured over a
population three times larger, because its four streams use the dictionary at
comparable breadth. Sections~\ref{sec:crit1} and~\ref{sec:crit2} report the pooled
medians; this decomposition is the reason for the gap.

\input{tab_modern_mid}

\section{Latent taxonomy and the window-size control}
\label{app:taxonomy}

Classifying every entry by which models' solo votes use it (Table~\ref{tab:taxonomy})
gives a shared core and an apparent model-specific tail. The tail depends on the window
size: of the 1,397 entries classified as specific on the 20k-token window, 305 remain
specific on the full 100k shard, 497 turn out shared, and 593 fall below the usage
threshold, while the consensus class is 92\% stable and grows with the window. The
entries that remain specific correlate at 0.84 with the other models' codes when the
joint TopK selects them; the label describes solo-vote ranking rather than
representational exclusivity. Causally, masking a model's own specific entries from its
solo vote changes its solo FVE by $\approx$0.001, while masking the consensus core
drives every model's solo FVE negative.

The correlations above are computed on codes after the joint TopK, so an entry the
selection never picks is unmeasured rather than absent. A probe with no TopK selection
therefore tests whether the remaining specific entries are exclusive. We
recompute every entry's responses with no selection at all, from the raw encoder
pre-activations $\operatorname{ReLU}(z_s)$ over the held-out window, and examine the 66
specific entries whose correlation under the joint TopK is below 0.2 or undefined. For each, we ask whether the
other three models respond on the owner's 50 strongest raw positions, calibrated
against two control groups: for consensus entries the best other model responds on a
median 100\% of the owner's top positions, and for well-correlated specific entries on
76\%. Of the 66 candidates, 41 turn out shared once the selection is removed, and one
responds weakly. The remaining 24 entries, 0.07\% of the dictionary, respond in one
model only, with a best other-model response rate of median 0\% and maximum 4\%;
21 of the 24 belong to Gemma, the extreme-scale stream.

\input{tab_modern_taxonomy}

\paragraph{Null and superseded checks.}
Four analyses were run and set aside. Identity-dominance of the assignment saturates
for every shared-selection system, so it cannot separate them. A magnitude comparison
restricted to co-support positions showed no signal beyond the value correlation. An
earlier solo-$\phi$ firing statistic is superseded by the solo value correlation of
Figure~\ref{fig:inversion}, which asks the same question directly under the solo
TopK, the deployment condition. We also rejected P(all)/P(any) firing ratios, because they reward the
selection mechanism itself rather than the latents.

\section{Planted-ground-truth check}
\label{app:planted}

The criteria of Section~\ref{sec:align} are computed on real activations, where the
true concepts are unknown, so we repeat the comparison on synthetic data with known
ground truth \citep{elhage2022toy,synthsaebench}. We plant 512 concepts and generate
three synthetic models of widths 256/192/128, five seeds each; injected noise caps the
attainable FVE. If forcing one shared concept-to-latent assignment cost anything, it
should recover fewer planted concepts than matching each model's SAE freely after
training. It does not: the shared assignment recovers the planted concepts exactly as
well as free per-model matching, at a universality ratio of 1.000, while recovery
collapses for a no-sharing control. Pooling across models also helps: the shared
dictionary recovers the planted concepts better than the per-model SAEs, with the
largest gain on the narrowest model.

\section{Shared rare latents: examples}
\label{app:examples}

Table~\ref{tab:examples} decodes the top activations of rare entries, selected on
20-99 of 100k held-out tokens, whose top-15 activating positions coincide 15 of 15
across all four models. Each shared latent decodes to the same word or construction
under four different tokenizers. Among the examples, two separate ``United'' detectors
give a clean example of feature splitting.

To make the reuse concrete, walk one row end to end. Latent 8319 fires 50 times per
100k held-out tokens, and it entered the table on firing agreement alone: its top-15
activating positions are the same 15 text positions in every model. Decoding those
positions yields the token ``free'' in each model's own tokenizer. A labeller can
therefore read the latent out of one model, record ``free'' detector once, and the
label is already correct for the other three models; a model adapted onto the frozen
dictionary that fires latent 8319 inherits the same meaning with no further work
(Section~\ref{sec:adapt}).

\input{tab_modern_examples}

Table~\ref{tab:semantic} gives the quantitative version of the criterion at the 4M
protocol: token-level decoded agreement and per-concept AP retention, for \ours{}
and the matched-dedicated control (Section~\ref{sec:crit3}).

\input{tab_modern_semantic}

\input{semanticity_appendix}

\section{Use of AI assistants}
\label{app:aiuse}

AI assistants were used throughout this project, in three roles. They drafted and
edited parts of the manuscript, including prose in the appendices. They wrote analysis
and figure code, and helped run the experiments on our cluster. They were used for
literature search, to locate related work we then read.

The authors take full responsibility for the content. Every number in the paper is
produced by code the authors reviewed and is read directly from the result files of the
runs described in Appendices~\ref{app:licenses} and~\ref{app:protocol}; no value is
quoted from an assistant's summary. Every citation was checked against the primary
source, arXiv, the ACL Anthology, or the publisher's page, and each was confirmed to
exist and to support the claim it is attached to. The assistants are tools here and are
not authors: they proposed no research question, and every claim in the paper was
verified by the authors before it was made.

\section{Cross-model probes: corpus transfer, causal effects, and steering}
\label{app:xmodel}

\paragraph{Sharedness off the training corpus.}
All sharedness numbers in Section~\ref{sec:align} are measured on FineWeb-Edu, the corpus
every model in the roster read during \ours{} training. To separate model-induced from
corpus-induced sharedness, we rebuild a 100k-position aligned store on Wikipedia with the
identical pipeline and re-measure the solo-TopK criteria with the converged dictionary,
which never saw this corpus. The picture is unchanged: the per-latent value-correlation
median is 0.72 over the 23,224 latents alive there (0.76 on the 12,510-latent
well-sampled core), and the four-way top-50 overlap median is 21 of 50. Sharedness is a
property of the models, not of the shared training text.

\paragraph{Scaled label agreement.}
We mine 250 contextual candidate latents (selected by usage breadth, solo value
correlation, and low token purity, i.e.\ not single-token detectors) plus 30 detectors,
and label a stratified 90 of them blind: for each latent, a label is written from each
model's own top activations separately, and the four labels are only then compared.
88 of 90 latents receive the same concept from all four models (2 partial, 0
disagreements). By level of abstraction, 24 are lexical, 51 phrasal or syntactic, and 15
represent abstract topics or discourse patterns.

\input{fig_fx_agreement}

\paragraph{Same latent, same causal effect.}
Figure~\ref{fig:fxagree} carries the distributions.
Beyond firing together, we test whether the same latent \emph{does} the same thing.
For each candidate latent and model we add $4\times$ its typical strong activation of the
model's own decoder direction to the hook layer over fresh text and rank tokens by the
induced mean logit change; agreement between two models is the overlap of their top-30
causally promoted token strings. The median over candidates is 0.19 (0.27 for the linear
logit-lens variant), against 0.00 under a mismatched-latent null. Both are lower bounds:
the overlap is computed on decoded strings, and the same concept often surfaces in
different surface forms (the geological-stratigraphy latent causally promotes
\emph{sedimentary, stratigraphic} in Gemma-3 and their Chinese equivalents in Qwen3),
while failure cases are real as well: subword-fragment features promote incoherent token
sets in every model.

\paragraph{Cross-model steering.}
We steer 12 labelled shared latents (10 high-level, 2 lexical detectors) in each model by
adding $\alpha$ times the latent's strong natural activation (the median of its top-20 solo
code values in that model) of the model's own unit decoder column to the hook layer at every
position while sampling 60-token continuations of 16 fixed prompts. A blind LLM jury scores
concept presence and coherence (1--5) over shuffled texts with conditions hidden; every
latent has a matched-norm random-latent control. Doses $\alpha \in \{0.5, 1\}$ use
repetition-penalized sampling and $\{2, 4\}$ plain sampling, so comparisons are made within
a dose regime (Figure~\ref{fig:steerdose}).

The dose--response is monotone in every model: the mean concept rate rises by 0.16--0.22
at $\alpha{=}1$ with coherence essentially preserved (3.0--3.3 against a 3.1--3.6 unsteered
baseline), by 0.30--0.42 at $\alpha{=}2$ at a visible coherence cost (2.2--2.7 against
3.8--4.3), and by roughly 0.5 at $\alpha{=}4$, where fluency collapses. The matched-norm
random controls stay within 0.04 of baseline at every dose: the movement is a property of
the labelled direction, not of perturbation magnitude. The three same-width streams move
nearly in lockstep: one dose scale transfers across models.

The failures are systematic rather than random. Topic-level features steer broadly at
$\alpha{=}2$ (geological stratigraphy and ALL-CAPS style in all four models; religious
discourse, war reporting, and nutrition in three), while token-level detectors and
positional slot features never steer at any dose (month names, large quantifiers,
citation years: zero successes), consistent with the input/output feature distinction
of \citet{arad-etal-2025-saes}: a feature that detects a token is not thereby a feature
that produces one. Gemma-3's sampler yields low judged coherence at every arm including
unsteered (1.0--1.8), so coherence-gated claims exclude it; its concept lift matches the
roster ($+0.22$ at $\alpha{=}1$). "Cross-model steering is therefore a trade-off curve rather than a single success rate.

\input{fig_steer_dose}

\input{app_hunt_examples}

%% file: fig_curves.tex
\begin{figure*}[t]
\centering
\includegraphics[width=0.32\textwidth]{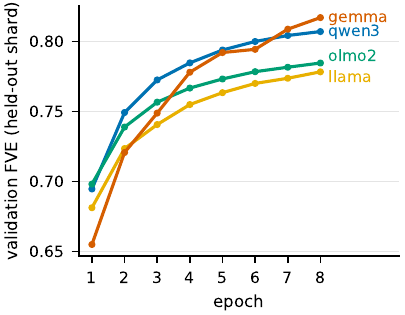}\hfill
\includegraphics[width=0.32\textwidth]{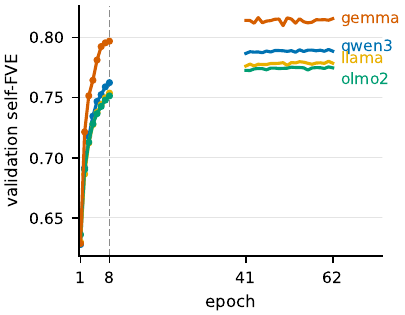}\hfill
\includegraphics[width=0.32\textwidth]{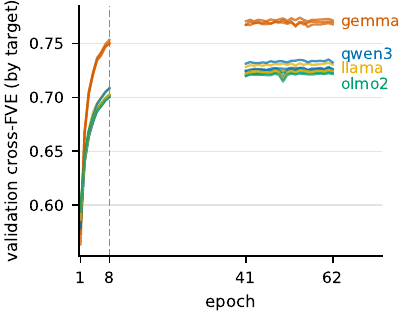}
\caption{Validation FVE by epoch at the 4M protocol: dedicated SAEs (left),
\ours{} self-reconstruction (middle), and \ours{} cross-reconstruction
(right, colored by target). Every curve rises monotonically through epoch 8
(dashed line), so the dedicated
ceilings and \ours{} train on to convergence (Section~\ref{sec:setup}). The middle
and right panels add the converged \ours{} run's logged tail, epochs 41-62, flat at
its early stop; a node requeue truncated that run's earlier log, and the probe
stream drawn for epochs 1-8 is a separate run (Appendix~\ref{app:metrics}).}
\label{fig:curves}
\end{figure*}

%% file: tab_modern_ablate.tex
\begin{table}[t]
\centering
\footnotesize
\setlength{\tabcolsep}{3pt}
\begin{tabular}{lccc}
\toprule
\multicolumn{4}{l}{\emph{Vote input (10 epochs, no dropout)}} \\
 & Share & \multicolumn{2}{c}{Vote shares (Q/L/O/G)} \\
\midrule
raw votes & 0.919 & \multicolumn{2}{c}{0.07/0.00/0.00/0.93} \\
RMS-cal.\ inputs & 0.925 & \multicolumn{2}{c}{0.24/0.40/0.30/0.07} \\
vote norm (raw) & \textbf{0.930} & \multicolumn{2}{c}{0.25/0.25/0.25/0.25} \\
\midrule
\multicolumn{4}{l}{\emph{Vote-source dropout dose (vote norm)}} \\
 & Share & Cross-FVE & Gemma solo \\
\midrule
$p=0$ & 0.930 & 0.595 & 0.155 \\
$p=0.25$ & 0.944 & \textbf{0.602} & 0.713 \\
$p=0.5$ & \textbf{0.945} & \textbf{0.602} & \textbf{0.727} \\
\midrule
\multicolumn{4}{l}{\emph{Cross-magnitude rescale (vote norm, dropout $p{=}0.5$)}} \\
 & Share & Cross-FVE & Gemma share \\
\midrule
oracle per-token & \textbf{0.945} & \textbf{0.602} & \textbf{0.920} \\
learned $c_m$ & 0.906 & 0.545 & 0.790 \\
no rescale & -3.034 & -42.096 & -11.715 \\
\bottomrule
\end{tabular}

\vspace{4pt}
\begin{tabular}{lcccc}
\toprule
\multicolumn{5}{l}{\emph{Cross-loss weight (10 epochs)}} \\
 & Share & Cross-FVE & Corr. & Top-50 \\
\midrule
$\lambda_x=1$ (settled) & \textbf{0.945} & \textbf{0.602} & \textbf{0.874} & \textbf{20/50} \\
$\lambda_x=0$ & 0.943 & 0.042 & 0.698 & 10/50 \\
$\lambda_x=0.5$ (no dropout) & 0.941 & 0.589 & - & - \\
\bottomrule
\end{tabular}
\caption{Ablations against the settled configuration (vote norm, vote-source dropout $p{=}0.5$,
oracle per-token cross rescale, $\lambda_x{=}1$); each block varies one ingredient with
the rest held fixed. Share, vote shares (Q/L/O/G), Corr., and Top-50 are defined in
Appendix~\ref{app:metrics}. Within each block, the best value per column is in bold.
Measured at the earlier protocol (800k-position store, $k{=}256$, 10 epochs, held-out shard~3), not rerun at the 4M protocol.}
\label{tab:ablate}
\end{table}

%% file: tab_modern_functional.tex
\begin{table*}[t]
\centering
\small
\setlength{\tabcolsep}{4.5pt}
\begin{tabular}{lcccccccc}
\toprule
 & \multicolumn{4}{c}{CE-recovered $\uparrow$} & \multicolumn{4}{c}{$\mathrm{KL}(\text{clean}\,\|\,\text{patched})$ $\downarrow$} \\
\cmidrule(lr){2-5}\cmidrule(lr){6-9}
 & Qwen3 & Llama & OLMo-2 & Gemma & Qwen3 & Llama & OLMo-2 & Gemma \\
\midrule
Dedicated (8-epoch) & \textbf{0.966} & 0.922 & 0.965 & 0.943 & \textbf{0.243} & 0.484 & 0.309 & 0.423 \\
\ours{} (settled) & 0.958 & \textbf{0.933} & \textbf{0.970} & \textbf{0.968} & 0.309 & \textbf{0.438} & \textbf{0.279} & \textbf{0.293} \\
\bottomrule
\end{tabular}
\caption{Functional splice at the hook layer under the solo TopK: the reconstruction
replaces the residual stream and next-token behavior is scored, per stream, against
the budget-matched 8-epoch dedicated SAEs (the converged dedicated splice is in Table~\ref{tab:main}). CE-recovered and KL are defined in
Appendix~\ref{app:metrics}. Best value per column in bold (higher CE-recovered, lower
KL).}
\label{tab:functional}
\end{table*}

%% file: tab_modern_consensus.tex
\begin{table*}[t]
\centering
\small
\begin{tabular}{lccccc}
\toprule
 & Qwen3 & Llama-3.2 & OLMo-2 & Gemma-3 & Mean \\
\midrule
\multicolumn{6}{l}{\emph{\ours{} (mean consensus size $j = 58.9$ of 128)}} \\
all own 128 latents & 0.804 & 0.794 & 0.795 & 0.821 & 0.803 \\
consensus latents only & 0.736 & 0.721 & 0.727 & 0.716 & 0.725 \\
own strongest $j$ latents & 0.758 & 0.740 & 0.746 & 0.739 & 0.746 \\
latents outside the consensus & -1.139 & -0.010 & 0.032 & -8.247 & -2.341 \\
\midrule
\multicolumn{6}{l}{\emph{USAE, 32 epochs (mean consensus size $j = 49.5$ of 128)}} \\
all own 128 latents & 0.719 & 0.708 & 0.717 & 0.756 & 0.725 \\
consensus latents only & 0.592 & 0.594 & 0.608 & 0.650 & 0.611 \\
own strongest $j$ latents & 0.649 & 0.638 & 0.650 & 0.682 & 0.655 \\
latents outside the consensus & 0.278 & 0.260 & 0.256 & 0.299 & 0.273 \\
\midrule
\multicolumn{6}{l}{\emph{SPARC (mean consensus size $j = 49.2$ of 128)}} \\
all own 128 latents & 0.727 & 0.701 & 0.697 & 0.136 & 0.565 \\
consensus latents only & 0.609 & 0.648 & 0.654 & -0.206 & 0.426 \\
own strongest $j$ latents & 0.636 & 0.661 & 0.664 & -0.152 & 0.452 \\
latents outside the consensus & -0.007 & 0.097 & 0.101 & -1.871 & -0.420 \\
\bottomrule
\end{tabular}
\caption{Reconstruction (FVE) through the consensus latents. For each token, each model's encoder alone selects
its top-128 latents under the solo TopK; the consensus set is the intersection of the
four selections. Each model is then
reconstructed from its own code restricted to: all 128 own
selections; the consensus set only; its own strongest $j$ latents, a size-matched
control; and its selections outside the consensus.}
\label{tab:consensus}
\end{table*}

%% file: tab_modern_overlap.tex
\begin{table}[t]
\centering
\footnotesize
\setlength{\tabcolsep}{2pt}
\begin{tabular}{lcc}
\toprule
\multicolumn{3}{l}{\emph{100k window, latents alive $\geq$ 10 times}} \\
 & Pairwise (of 50) & All-four \\
System & median (span) & (of 50) \\
\midrule
\ours{} & \textbf{37} (37-38) & \textbf{26} \\
Matched dedicated & 15 (11-17) & - \\
USAE (8 epochs) & 0 (0-0) & - \\
\midrule
\multicolumn{3}{l}{\emph{300k window, latents with $\geq$ 50 firings ($n$)}} \\
 & Joint TopK & Solo TopK \\
\midrule
\ours{} & 33.5 (19,211) & \textbf{31.5} (21,123) \\
SPARC & \textbf{34} (7,054) & 0 (28,982) \\
USAE (32 epochs) & -  & 28 (13,065) \\
Matched dedicated & -  & 13.5 (31,742) \\
\bottomrule
\end{tabular}
\caption{Top-50 activating-position overlap: how many of a latent's 50 top-activating
positions coincide across models. Top block: median over the six model pairs (span in
parentheses) and the all-four count on the core of latents alive at least 50 times in
every model. Bottom block: the same statistic under each selection condition on the 300k window, with each
system's latent population in parentheses; USAE codes are per-model native, so only the
solo column is defined. Chance overlap is 0 of 50. Dashes: statistic undefined or
not computed for the system (Appendix~\ref{app:metrics}). Best value per column in
bold.}
\label{tab:overlap}
\end{table}

%% file: tab_modern_features.tex
\begin{table}[t]
\centering
\footnotesize
\setlength{\tabcolsep}{3.5pt}
\begin{tabular}{lrr}
\toprule
\multicolumn{3}{l}{\emph{Vocabulary classes ($\tau=0.1\%$ usage, 100k-token window)}} \\
Class & Latents & Med.\ corr. \\
\midrule
Consensus (all four) & 10,776 & 0.91 \\
Partial (two or three) & 1,650 & 0.85 \\
Specific (one model) & 1,144 & 0.84 \\
Below $\tau$ everywhere & 19,190 & 0.82 \\
\midrule
\multicolumn{3}{l}{\emph{Stability of the 20k-window classes at 100k}} \\
\multicolumn{2}{l}{Consensus staying consensus} & 92\% \\
\multicolumn{2}{l}{``Specific'' staying specific} & 22\% \\
\multicolumn{2}{l}{``Specific'' moving up to shared} & 36\% \\
\multicolumn{2}{l}{``Specific'' falling below $\tau$} & 42\% \\
\midrule
\multicolumn{3}{l}{\emph{Per-stream statistics (Qwen3/Llama/OLMo-2/Gemma)}} \\
\multicolumn{2}{l}{Median token purity} & 0.12/0.12/0.12/0.12 \\
\multicolumn{2}{l}{Single-token detectors (\%)} & 1.4/1.3/1.6/1.7 \\
\multicolumn{2}{l}{Solo-joint code corr.\ (median)} & 0.96/0.96/0.95/0.94 \\
\multicolumn{2}{l}{Solo-joint vocabulary overlap} & 0.92/0.91/0.90/0.90 \\
\multicolumn{2}{l}{Solo-vocabulary Jaccard (all pairs)} & 0.84-0.86 \\
\bottomrule
\end{tabular}
\caption{Solo-TopK usage classes at $\tau=0.1\%$ of tokens, the migration of the
20k-window classes when the window grows to 100k tokens, and per-stream statistics.
\emph{Med.\ corr.}\ is the median per-latent cross-model code correlation within the
class; populations and definitions in Appendix~\ref{app:metrics}.}
\label{tab:features}
\end{table}

%% file: fig_convergence.tex
\begin{figure}[t]
\centering
\includegraphics[width=\columnwidth]{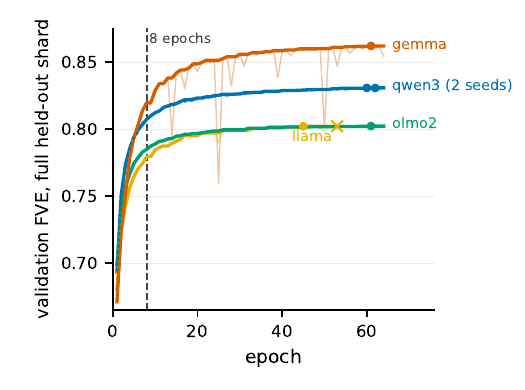}
\caption{Dedicated ceilings trained until validation FVE levels off, under the
constant-rate recipe and a 64-epoch cap. Bold: the best validation FVE reached so
far - the checkpointed ceiling; faint: the per-epoch series. Dots mark each
stream's peak; the cross marks Llama's early stop (peak unbeaten for eight
consecutive epochs); the dashed line is the 8-epoch budget of every shared
system. The second Qwen3 seed (dotted) is indistinguishable from the first
throughout. This full-shard window sits 0.5-1.0 points of FVE below the evaluation
window of Table~\ref{tab:main} (Appendix~\ref{app:metrics}).}
\label{fig:convergence}
\end{figure}

%% file: fig_yardstick.tex
\begin{figure}[t]
\centering
\includegraphics[width=\columnwidth]{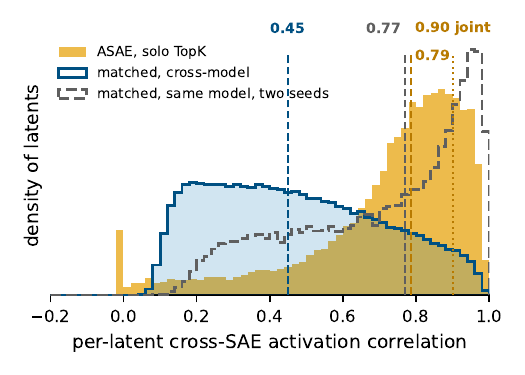}
\caption{The post-hoc-matching reference for the second criterion. Per-latent
activation correlations: the best post-hoc match between separately trained dedicated
SAEs of \emph{different} models (blue; converged arms, median 0.45), the same matching
between two SAEs of the \emph{same} model trained from different seeds (gray; median
0.77) - the ceiling the post-hoc route permits with no cross-model disagreement at
all - and \ours{}'s shared-latent correlation with each model voting alone (orange;
median 0.79; the dotted tick marks the joint-TopK median, 0.90). Matching cannot reach
its own within-model ceiling across models; the shared dictionary clears that ceiling
solo and exceeds it further under the joint vote. \ours{}'s near-zero mass is rare
latents - 3\% of measured latents, 0.1\% of solo selections (Section~\ref{sec:crit2}). Matching controls are measured on the 100k
window and \ours{} correlations on the 300k window; the within-model arm is one seed
pair on one stream (protocols in Appendix~\ref{app:metrics}).}
\label{fig:yardstick}
\end{figure}

%% file: fig_loo_sharedness.tex
\begin{figure}[t]
\centering
\includegraphics[width=\columnwidth]{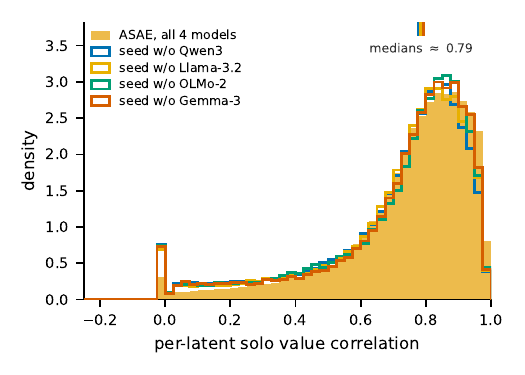}
\caption{Solo per-latent value-correlation distributions inside each three-model
leave-one-out seed (step outlines, colored by the \emph{held-out} model) over the
full four-model \ours{} (orange fill). The four seed medians coincide with each other
and with the full dictionary's.}
\label{fig:loo}
\end{figure}

%% file: tab_modern_loo.tex
\begin{table}[t]
\centering\scriptsize
\setlength{\tabcolsep}{2.6pt}
\begin{tabular}{lccccc}
\toprule
seed & corr. & top-50 & pairwise & $N$-way & FVE/conv. \\
\midrule
without Qwen3-1.7B & 0.778 & 32.0 & 83.3 & 71.8 (3-way) & 0.945 \\
without Llama-3.2-1B & 0.784 & 32.0 & 85.9 & 75.0 (3-way) & 0.942 \\
without OLMo-2-1B & 0.791 & 32.0 & 85.8 & 74.8 (3-way) & 0.935 \\
without Gemma-3-1B & 0.792 & 32.0 & 80.9 & 68.5 (3-way) & 0.956 \\
full \ours{} & 0.797 & 31.5 & 78.0 & 59.0 (4-way) & 0.967 \\
\bottomrule
\end{tabular}
\caption{Internal sharedness of each three-model leave-one-out seed, members voting
solo: pooled per-latent value-correlation median and top-50 overlap median (300k
window), mean pairwise and $N$-way per-token selection intersection of 128 (e02
window), and mean solo FVE as a share of the members' converged dedicated ceilings.
The seeds train for 8 epochs, so their shares sit below the converged
four-model row by the ceiling gap, not by any property of the roster. The full-model
row restates published values.}
\label{tab:loo}
\end{table}

%% file: tab_modern_sparcsolo.tex
\begin{table*}[t]
\centering\scriptsize
\begin{minipage}[t]{0.53\textwidth}
\centering
\begin{tabular}{lcccc}
\toprule
under the solo TopK & Qwen3 & Llama & OLMo & Gemma \\
\midrule
SPARC, vote share & 0.245 & 0.276 & 0.274 & 0.205 \\
SPARC, solo $\cap$ joint & 0.711 & 0.625 & 0.592 & 0.635 \\
SPARC, latents used & 22,933 & 30,635 & 31,097 & 7,719 \\
\ours{}, latents used & 24,328 & 24,776 & 24,532 & 24,964 \\
\midrule
\multicolumn{5}{l}{\emph{Four-way intersection of the solo selections, of 128}} \\
SPARC 49.2 \quad \ours{} 58.9
& & & & \\
\bottomrule
\end{tabular}
\end{minipage}\hfill
\begin{minipage}[t]{0.44\textwidth}
\centering
\begin{tabular}{lccc}
\toprule
 & per token & \multicolumn{2}{c}{per latent ($n$ measurable)} \\
\cmidrule(lr){2-2}\cmidrule(lr){3-4}
SPARC pair & of 128 & top-50 & value corr. \\
\midrule
Qwen3-Llama & 67.4 & 0 (15,233) & 0.00 (24,959) \\
Qwen3-OLMo & 63.8 & 0 (15,514) & 0.00 (25,029) \\
Llama-OLMo & 59.7 & 0 (27,559) & 0.00 (31,395) \\
Qwen3-Gemma & 68.5 & 34 (7,092) & 0.77 (7,313) \\
Llama-Gemma & 62.0 & 33 (7,232) & 0.75 (7,578) \\
OLMo-Gemma & 59.2 & 32 (7,248) & 0.74 (7,590) \\
\bottomrule
\end{tabular}
\end{minipage}
\caption{SPARC's solo-TopK condition. Left: selection behavior on the
scoring window of Table~\ref{tab:main} - vote share is each stream's mean share of
the aggregated selection mass, solo $\cap$ joint the fraction of the joint top-$k$
that a stream's own top-$k$ reproduces, and latents used the number of distinct
latents a stream's solo vote ever selects (the joint selection uses
7,599 for SPARC and 23,714 for \ours{}). Right: the same
condition per model pair - how many of the 128 selections the two streams share at a
token, and the two per-latent medians of Table~\ref{tab:overlap} and
Figure~\ref{fig:inversion} with the population they are measured over (300k window).
Our joint selection was verified to reproduce SPARC's own \texttt{shared\_indices}
exactly on every batch.}
\label{tab:sparcsolo}
\end{table*}

%% file: tab_modern_mid.tex
\begin{table}[t]
\centering\footnotesize
\setlength{\tabcolsep}{3pt}
\begin{tabular}{lcccccc}
\toprule
model & ded. & joint & share & solo & solo/ded & sink\,\% \\
\midrule
Qwen3 & 0.807 & 0.768 & 0.952 & 0.767 & 0.951 & 0.015 \\
Llama-3.2 & 0.782 & 0.738 & 0.944 & 0.753 & 0.963 & 0.000 \\
OLMo-2 & 0.712 & 0.653 & 0.917 & 0.681 & 0.957 & 0.000 \\
Gemma-3 & 0.819 & 0.782 & 0.954 & 0.751 & 0.917 & 0.000 \\
\bottomrule
\end{tabular}
\caption{Mid-depth (\(\sim\)50\%) roster: the settled \ours{} (joint and solo TopK)
against dedicated (ded.) ceilings at hooks qwen3-14 llama-8 olmo2-8 gemma-13 (mid \~{}50\%). share and
solo/ded are ratios to the dedicated FVE, and sink\,\% is defined in
Appendix~\ref{app:metrics}. Cross-FVE mean is 0.652 (max 0.743).
Measured at the earlier protocol (800k-position store, $k{=}256$, 10 epochs, held-out shard~3), not rerun at the 4M protocol.}
\label{tab:mid}
\end{table}

%% file: tab_modern_taxonomy.tex
\begin{table}[t]
\centering\footnotesize
\setlength{\tabcolsep}{3.5pt}
\begin{tabular}{lccc}
\toprule
class & @20k & @100k & corr \\
\midrule
consensus (all four) & 9,987 & 10,776 & 0.91 \\
partial (2-3 models) & 1,877 & 1,650 & 0.85 \\
specific (one model) & 1,397 & 1,144 & 0.84 \\
joint-only & 17 & 8 & 0.72 \\
below $\tau$ everywhere & 19,490 & 19,190 & 0.82 \\
\bottomrule
\end{tabular}
\caption{Latent classes by solo-TopK usage (threshold $\tau=0.1\%$ of tokens):
latent counts when classified on a 20k-token window vs.\ the full 100k held-out shard;
``corr'' is the class's median cross-model code correlation at 100k. The migration
between the two windows is quantified in Table~\ref{tab:features} and
Appendix~\ref{app:taxonomy}.}
\label{tab:taxonomy}
\end{table}

%% file: tab_modern_examples.tex
\begin{table*}[t]
\centering\small
\begin{tabular}{llcccc}
\toprule
latent & fires/100k & Qwen3-1.7B & Llama-3.2-1B & OLMo-2-1B & Gemma-3-1B \\
\midrule
8319 & 50 & \texttt{free} & \texttt{free} & \texttt{free} & \texttt{free} \\
29780 & 52 & \texttt{i} & \texttt{i} & \texttt{i} & \texttt{i} \\
11723 & 75 & \texttt{,'', ,"} & \texttt{,", ,''} & \texttt{,'', ,"} & \texttt{,", ,''} \\
16628 & 22 & \texttt{United} & \texttt{United} & \texttt{United} & \texttt{United} \\
30588 & 20 & \texttt{United} & \texttt{United} & \texttt{United} & \texttt{United} \\
10904 & 56 & \texttt{bar, Bar} & \texttt{bar, Bar, -bar} & \texttt{bar, Bar} & \texttt{bar, Bar} \\
\bottomrule
\end{tabular}
\caption{Rare shared latents: latents selected on 20-99 of 100k held-out tokens whose
top-15 activating positions coincide across all four models. Cells show each model's top
activating tokens, decoded by its own tokenizer. Six of the eight sampled latents are
shown; both ``United'' detectors are kept, an instance of feature splitting.
Measured at the earlier protocol (800k-position store, $k{=}256$, 10 epochs, held-out shard~3), not rerun at the 4M protocol.}
\label{tab:examples}
\end{table*}

%% file: tab_modern_semantic.tex
\begin{table}[t]
\centering
\footnotesize
\setlength{\tabcolsep}{1.5pt}
\begin{tabular}{lccc}
\toprule
\multicolumn{4}{l}{\emph{Token level: ten strongest positions decoded in}} \\
\multicolumn{4}{l}{\emph{each model's own tokenizer (medians over 6 pairs)}} \\
 &  & \ours{} & Matched ded. \\
\midrule
\multicolumn{2}{l}{Modal decoded-token agreement} & \textbf{0.63} & 0.45 \\
\multicolumn{2}{l}{Decoded token-set Jaccard} & \textbf{0.50} & 0.19 \\
\midrule
\multicolumn{4}{l}{\emph{Concept level: AP retention of each model's}} \\
\multicolumn{4}{l}{\emph{five most selective detectors}} \\
 & AP in &  &  \\
Concept & own model & \ours{} & Matched ded. \\
\midrule
Months & 0.433 & 0.999 & \textbf{1.004} \\
Numbers & 0.411 & \textbf{1.000} & 0.783 \\
3rd-person pronouns & 0.273 & \textbf{1.000} & 0.909 \\
First person & 0.238 & 0.998 & \textbf{1.025} \\
Countries & 0.149 & \textbf{1.000} & 0.854 \\
Units & 0.183 & 0.990 & \textbf{1.036} \\
Reporting verbs & 0.229 & \textbf{1.000} & 0.998 \\
Time words & 0.338 & \textbf{1.000} & 0.814 \\
Family & 0.187 & \textbf{1.000} & 0.976 \\
Body & 0.198 & 1.000 & \textbf{1.015} \\
\midrule
Pooled ($n{=}600$/591) & -- & \textbf{1.000} & 0.973 \\
\bottomrule
\end{tabular}
\caption{Semantic readout under \ours{} and under the matched dedicated control, at
the token level and at the concept level. The AP-in-own-model column gives each
detector's base selectivity; retention is the ratio of a detector's average precision
read out in the receiving model to that base value. Protocol in
Appendix~\ref{app:metrics}. Bold: higher retention per row; ties at the printed
precision are both bolded.}
\label{tab:semantic}
\end{table}

%% file: semanticity_appendix.tex
\section{FADE Semantic Evaluation}
\label{app:fade}

We evaluate whether the shared latent dictionary admits model-independent semantic interpretations using the FADE framework~\citep{puri-etal-2025-fade}. FADE measures four properties of a semantic description: \emph{clarity}, \emph{responsiveness}, \emph{purity}, and \emph{faithfulness}. The protocol below describes label generation, filtering, and cross-model evaluation.

\paragraph{Evaluation dataset.}

FADE evaluation uses the same aligned corpus as \ours{} training. Latent labels are generated from one million aligned token positions, while FADE scores are computed on a disjoint held-out set of 200\,000 aligned positions. No examples used for label generation are included during evaluation.

\paragraph{Example retrieval and label generation.}

For each evaluated latent, we retrieve its top activating examples. We consider two labelling strategies.

\emph{Shared-vote labelling} selects examples according to the shared activation vote, and generates a single description from these examples. This label is attached to the shared latent and evaluated on every model.

\emph{Per-stream labelling} selects examples independently using each model's own latent activations, producing one label per model and latent. These labels are evaluated both on their source stream and transferred to the other streams.

For both strategies, the labeling model receives the retrieved textual contexts and is instructed to produce a concise semantic description capturing the common property responsible for the activation.

\paragraph{Latent sampling.}

We evaluate 100 latents sampled across activation-frequency strata: 20 high-frequency latents (activation rate $\geq 1\%$), 60 medium-frequency latents (between $0.1\%$ and $1\%$), and 20 low-frequency latents (below $0.1\%$). The same sampled latents are evaluated across all four models.

\paragraph{FADE evaluation.}

For each label, FADE scores are computed independently on each model using that model's encoder and TopK latent activations.

\begin{itemize}
    \item \textbf{Clarity} measures whether the description is specific and unambiguous.
    \item \textbf{Responsiveness} measures whether latent activations correspond to the described concept.
    \item \textbf{Purity} measures whether the description uniquely identifies the latent compared with unrelated activations.
    \item \textbf{Faithfulness} measures whether intervening on the latent causally affects the described behavior.
\end{itemize}

Because semantic quality is only meaningful for labels that correctly identify the latent, we use a clarity threshold of $0.9$ for selecting reliable labels.

\paragraph{Evaluation settings.}

We report three complementary evaluations.

\emph{Per-stream evaluation} measures the quality of labels on the stream where they were validated. A label is retained if its clarity is at least $0.9$ on the evaluated stream.

\emph{Cross-stream transfer evaluation} measures whether a label generated from one model remains valid on other models. For per-stream labels, a label is retained if it achieves clarity $\geq 0.9$ on its source stream, and the same description is then evaluated on all streams.

\emph{Cross-model consistency} measures whether semantic scores agree across architectures. For each latent, we compute FADE scores independently for every model and calculate Pearson correlations across the six model pairs. These correlations quantify whether latents that are easy (or difficult) to interpret in one model remain similarly interpretable in others.

\begin{table*}[t]
\centering
\footnotesize
\setlength{\tabcolsep}{3pt}
\renewcommand{\arraystretch}{1.15}
\begin{tabular}{
  r
  >{\centering\arraybackslash}p{0.15\textwidth}
  >{\centering\arraybackslash}p{0.17\textwidth}
  >{\centering\arraybackslash}p{0.14\textwidth}
  >{\centering\arraybackslash}p{0.15\textwidth}
  >{\centering\arraybackslash}p{0.17\textwidth}
}
\toprule
Latent & Per-stream Qwen3 & Per-stream OLMo-2 & Per-stream Gemma-3 & Per-stream Llama-3.2 & Shared Vote \\
\midrule
16291 & word `in' or `of' & word `of' or `to' or punctuation marks & word `of' or `in' & word `of' or `is' or `in' & prepositions and articles \\
2317 & four-year term or similar duration & four-year term or similar duration & year term election context & four-year term or similar duration & four-year term or similar duration \\
6418 & using or a before handheld gps & handheld gps usage & word-initial `a' or `using' & word-initial `a' or `the' & word-initial `a' or `using' \\
2727 & word-initial uppercase fragment & word-initial uppercase fragment & word-initial uppercase fragment & word-initial uppercase fragment & word-initial uppercase fragment \\
14390 & possessive form & possessive form & word-initial uppercase fragment & possessive `s or whose & possessive form \\
22504 & should & should & should fragment & should & should \\
5426 & word-initial `a' or `the' & word-initial `a' or `the' & word-initial `the' or `a' & word-initial `the' or `a' & the/called/a \\
17068 & marijuana or drug-related terms & food or drug-related terms & marijuana or drug & marijuana or drug & marijuana or drug-related terms \\
20781 & word-initial uppercase fragment & word-initial uppercase fragment & word-initial uppercase fragment & class or gourmet & word-initial uppercase fragment \\
\bottomrule
\end{tabular}
\caption{Examples of shared \ours{} latents and their per-stream semantic descriptions.}
\label{tab:shared_feature_examples}
\end{table*}

\begin{figure*}[t]
\centering
\includegraphics[width=\textwidth]{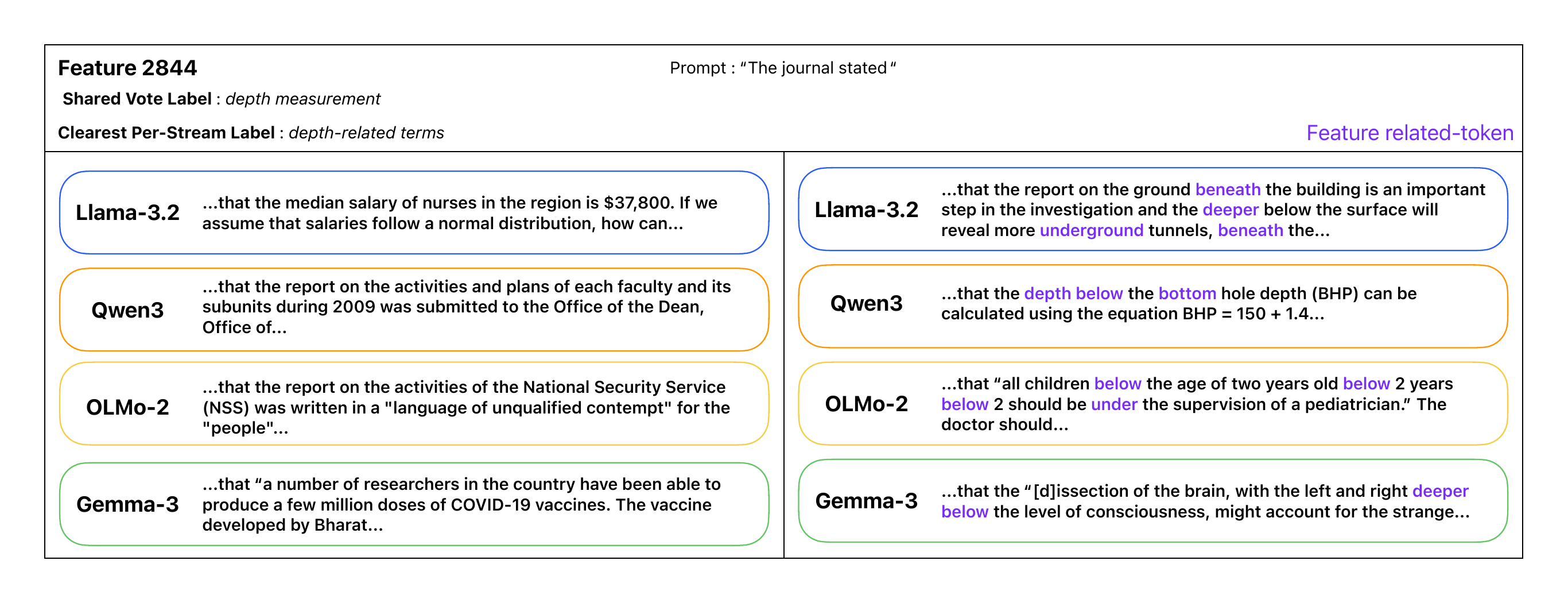}
\caption{Qualitative steering examples for shared \ours{} latents activated across all four models. The same shared latent is steered independently in each model, illustrating that semantically aligned latents can induce related behavioral shifts while leaving room for model-specific effects.}
\label{fig:steer_examples}
\end{figure*}

\begin{table}[t]
\centering
\small
\begin{tabular}{lcc}
\toprule
\textbf{Dedicated baseline} & \textbf{$t$} & \textbf{$p$} \\
\midrule
SAE Gemma-3 & -1.154 & 0.259 \\
SAE Llama-3.2 & -1.002 & 0.327 \\
SAE OLMo-2 & -0.210 & 0.840 \\
SAE Qwen3 & -0.420 & 0.679 \\
\bottomrule
\end{tabular}
\caption{Welch two-sample tests comparing FADE faithfulness between \ours{}
shared-vote labels and each dedicated SAE baseline on retained label--stream
pairs. None of the comparisons is significant at $p < 0.05$.}
\label{tab:fade_faithfulness_tests}
\end{table}

%% file: fig_fx_agreement.tex
\begin{figure}[t]
\centering
\includegraphics[width=\columnwidth]{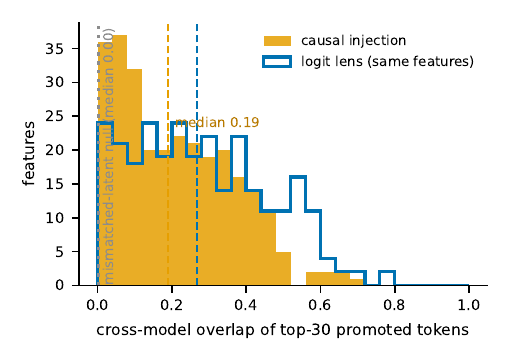}
\caption{Same latent, same effect: for each of the 280 mined candidate latents, the
overlap of the top-30 tokens the latent promotes across the four models (median over
the six pairs), measured causally by injection at the hook layer (filled) and by the
linear logit lens (outline). A mismatched-latent null sits at 0.00. String-level
overlap is a lower bound: the same concept often surfaces in different languages and
subword forms across tokenizers (Appendix~\ref{app:xmodel}).}
\label{fig:fxagree}
\end{figure}

%% file: fig_steer_dose.tex
\begin{figure}[t]
\centering
\includegraphics[width=\columnwidth]{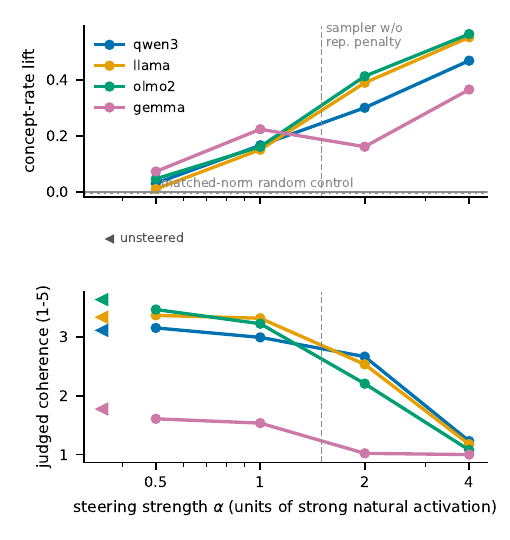}
\caption{Cross-model steering dose--response, mean over the 12 steered latents. Top:
concept-rate lift over each dose regime's own unsteered baseline; the dotted line is the
matched-norm random-latent control. Bottom: judged coherence; arrowheads mark unsteered
baselines. Doses left of the dashed line use repetition-penalized sampling, right of it
plain sampling. The three width-2048 streams move nearly in lockstep; Gemma-3's sampler
bounds its judged coherence at every arm.}
\label{fig:steerdose}
\end{figure}

%% file: app_hunt_examples.tex
\section{Top-activating contexts for the probe-found latents}
\label{app:huntex}

For each latent of Figure~\ref{tab:hunt}: per model, the examples at ranks 1, 5, and
10 of its own solo-vote activations on the held-out window - sampling depth rather
than the contiguous top, so the displayed texts diversify - each annotated with how
many models hold that exact position in their own top-10. Cross-model repetition of a
passage is the position-overlap statistic made visible, not duplicated evidence. Where the source document was recovered, the full original
sentence is shown with the activating word restored; \textdagger{} marks contexts still
rendered from span-final tokens (word-internal subwords absent).
Tables~\ref{tab:huntex1537}--\ref{tab:huntex30007}.

\begin{table*}[p]
\centering
\footnotesize
\begin{tabular}{p{0.96\textwidth}}
\toprule
\textbf{\#1537} - positive appraisal (English and French probes) \\
\midrule
\multicolumn{1}{l}{\emph{Qwen3}} \\
\quad {\scriptsize(r1, in 4/4 top-10)} \ldots ists. |The spine and ribs ofgnathodon saturator| It's a very\textbf{ nice} display. And while the amount  \ldots\textsuperscript{\textdagger} \\
\quad {\scriptsize(r5, in 4/4 top-10)} \ldots hem in a history that is both analytical and narrative...A\textbf{ superb} book that will be a standard m \ldots\textsuperscript{\textdagger} \\
\quad {\scriptsize(r10, in 4/4 top-10)} \ldots inifa should get an Israel Business award! What a Kiddushem!\textbf{ Nice} to hear about the positive asp \ldots\textsuperscript{\textdagger} \\
\multicolumn{1}{l}{\emph{Llama-3.2}} \\
\quad {\scriptsize(r1, in 4/4 top-10)} \ldots ists. |The spine and ribs ofgnathodon saturator| It's a very\textbf{ nice} display. And while the amount  \ldots\textsuperscript{\textdagger} \\
\quad {\scriptsize(r5, in 1/4 top-10)} \ldots ouses, and- don't forget to provide water. Pamigging has\textbf{ marvelous} shots of bees visiting her st \ldots\textsuperscript{\textdagger} \\
\quad {\scriptsize(r10, in 1/4 top-10)} It was Her wont, long before the birth of Her Son, to ponder in Her heart16 and evaluate all things with the inner eye of the heart. In Western Christianity we encounter the description of some people as interior souls. \\
\multicolumn{1}{l}{\emph{OLMo-2}} \\
\quad {\scriptsize(r1, in 4/4 top-10)} \ldots ists. |The spine and ribs ofgnathodon saturator| It's a very\textbf{ nice} display. And while the amount  \ldots\textsuperscript{\textdagger} \\
\quad {\scriptsize(r5, in 2/4 top-10)} \ldots gical Society (Photo credit: Jay Landolfi). Gordon is a\textbf{ magnificent} speaker but very seldom give \ldots\textsuperscript{\textdagger} \\
\quad {\scriptsize(r10, in 1/4 top-10)} \ldots gue fern (Asiumolendrium| erns can give you some really\textbf{ interesting} clues about the environment  \ldots\textsuperscript{\textdagger} \\
\multicolumn{1}{l}{\emph{Gemma-3}} \\
\quad {\scriptsize(r1, in 4/4 top-10)} \ldots ists. |The spine and ribs ofgnathodon saturator| It's a very\textbf{ nice} display. And while the amount  \ldots\textsuperscript{\textdagger} \\
\quad {\scriptsize(r5, in 2/4 top-10)} \ldots gical Society (Photo credit: Jay Landolfi). Gordon is a\textbf{ magnificent} speaker but very seldom give \ldots\textsuperscript{\textdagger} \\
\quad {\scriptsize(r10, in 3/4 top-10)} \ldots osals, Lewis D. Hopkins and Marisa A.ataPerhaps the most\textbf{ remarkable} thing about the image above  \ldots\textsuperscript{\textdagger} \\
\multicolumn{1}{l}{\emph{Qwen3.5-2B (adapted)}\hfill {\scriptsize value corr.\ with seeds 0.87--0.88}} \\
\quad \ldots visionUsing the0 Chart to Count MoneyThe other day I saw the\textbf{atest} way of using a0\textsuperscript{\textdagger} \\
\quad \ldots n Steel, The Early Years in Canada,3-5 by Craigon is an\textbf{ interesting} book and the quotes by\textsuperscript{\textdagger} \\
\quad (By the way, this is a \textbf{great} example to indicate that the biggest bias in the media is towards news, not any particular side of a story). \\
\multicolumn{1}{l}{\emph{Llama-3.2-1B-Instruct (adapted)}\hfill {\scriptsize value corr.\ with seeds 0.84--0.91}} \\
\quad The proprietor of the recently named ``Best Science Blog'', also had a couple of choice comments: In my opinion, this press release and subsequent media interviews were done for media attention. \\
\quad \ldots visionUsing the0 Chart to Count MoneyThe other day I saw the\textbf{atest} way of using a0\textsuperscript{\textdagger} \\
\quad \ldots n Steel, The Early Years in Canada,3-5 by Craigon is an\textbf{ interesting} book and the quotes by\textsuperscript{\textdagger} \\
\multicolumn{1}{l}{\emph{Hunyuan-1.8B (adapted, cross-lab)}\hfill {\scriptsize value corr.\ with seeds 0.87--0.88}} \\
\quad {\scriptsize(r1, in 4/4 seed top-10)} \ldots n Steel, The Early Years in Canada,3-5 by Craigon is an\textbf{ interesting} book and the quotes by theMA\textsuperscript{\textdagger} \\
\quad {\scriptsize(r5, in 3/4 seed top-10)} The proprietor of the recently named ``Best Science Blog'', also had a couple of choice comments: In my opinion, this press release and subsequent media interviews were done for media attention. \\
\quad {\scriptsize(r10, in 1/4 seed top-10)} \ldots  :local capture" by vested interests. There are several\textbf{ encouraging} points that emerge from thes \ldots\textsuperscript{\textdagger} \\
\bottomrule
\end{tabular}
\caption{Latent 1537: examples at activation ranks 1/5/10 per model under its own solo
vote; the activating word is bold; ($r$, in $k$/4 top-10) gives the example's rank and
how many models share the position. Top-50 activations span 38--43 distinct passages per model.}
\label{tab:huntex1537}
\end{table*}

\begin{table*}[p]
\centering
\footnotesize
\begin{tabular}{p{0.96\textwidth}}
\toprule
\textbf{\#7393} - general negative affect (moral-wrongdoing and distress probes) \\
\midrule
\multicolumn{1}{l}{\emph{Qwen3}} \\
\quad {\scriptsize(r1, in 4/4 top-10)} But the power it would need to guarantee that no one is poor would be so great it could crush the natural rights and liberty of individuals. \\
\quad {\scriptsize(r5, in 3/4 top-10)} Smederevo citizens, frightened and hurt by a \textbf{harsh} occupation, and on April 6, by the bombing of the city, too, withdrew to their homes and only came out on the streets when needed: to finish their jobs and to procure food. \\
\quad {\scriptsize(r10, in 3/4 top-10)} Those disciples of Christ who have experienced their status to any degree have become keenly aware of the intimate relationship they have with Marya relationship that at the moment is perhaps closer than that with Christ. For it is Mary who brings us to Christ. \\
\multicolumn{1}{l}{\emph{Llama-3.2}} \\
\quad {\scriptsize(r1, in 4/4 top-10)} He concludes-echoing historian Rick Atkinson's excellent recent account of the campaign, The Day of Battle-that despite its \textbf{terrible} cost, the fight in Italy played a decisive role in defeating Germany. \\
\quad {\scriptsize(r5, in 4/4 top-10)} \ldots  scale.However all these words seen apt at describing the\textbf{ terrible} afflictions of cancer thatppe \ldots\textsuperscript{\textdagger} \\
\quad {\scriptsize(r10, in 3/4 top-10)} Those disciples of Christ who have experienced their status to any degree have become keenly aware of the intimate relationship they have with Marya relationship that at the moment is perhaps closer than that with Christ. For it is Mary who brings us to Christ. \\
\multicolumn{1}{l}{\emph{OLMo-2}} \\
\quad {\scriptsize(r1, in 4/4 top-10)} He concludes-echoing historian Rick Atkinson's excellent recent account of the campaign, The Day of Battle-that despite its \textbf{terrible} cost, the fight in Italy played a decisive role in defeating Germany. \\
\quad {\scriptsize(r5, in 3/4 top-10)} Many of them, especially students from the village were, at the time of explosion of munitions, at the train station and on the trains to return to their villages, so, unfortunately, they found \textbf{terrible} death and harrowing! \\
\quad {\scriptsize(r10, in 3/4 top-10)} Those disciples of Christ who have experienced their status to any degree have become keenly aware of the intimate relationship they have with Marya relationship that at the moment is perhaps closer than that with Christ. For it is Mary who brings us to Christ. \\
\multicolumn{1}{l}{\emph{Gemma-3}} \\
\quad {\scriptsize(r1, in 4/4 top-10)} But the power it would need to guarantee that no one is poor would be so great it could crush the natural rights and liberty of individuals. \\
\quad {\scriptsize(r5, in 3/4 top-10)} Many of them, especially students from the village were, at the time of explosion of munitions, at the train station and on the trains to return to their villages, so, unfortunately, they found \textbf{terrible} death and harrowing! \\
\quad {\scriptsize(r10, in 3/4 top-10)} For the Right wing community to feel that they are not under such harsh accusation, they too can now begin to grapple with aspects of this \textbf{terrible} crime. \\
\multicolumn{1}{l}{\emph{Qwen3.5-2B (adapted)}\hfill {\scriptsize value corr.\ with seeds 0.80--0.85}} \\
\quad But the power it would need to guarantee that no one is poor would be so great it could crush the natural rights and liberty of individuals. \\
\quad \ldots overnment brought down on their heads.shima signifies the ug\textbf{liest} dimension of all this.\textsuperscript{\textdagger} \\
\quad It is quite possible that few Japanese care about the atrocities Japan committed overseas. At the same time, many Japanese do care about the Japanese victims, and about the \textbf{grotesque} violence their militarist government brought down on their heads. \\
\multicolumn{1}{l}{\emph{Llama-3.2-1B-Instruct (adapted)}\hfill {\scriptsize value corr.\ with seeds 0.79--0.89}} \\
\quad It is quite possible that few Japanese care about the atrocities Japan committed overseas. At the same time, many Japanese do care about the Japanese victims, and about the \textbf{grotesque} violence their militarist government brought down on their heads. \\
\quad But the power it would need to guarantee that no one is poor would be so great it could crush the natural rights and liberty of individuals. \\
\quad \ldots overnment brought down on their heads.shima signifies the ug\textbf{liest} dimension of all this.\textsuperscript{\textdagger} \\
\multicolumn{1}{l}{\emph{Hunyuan-1.8B (adapted, cross-lab)}\hfill {\scriptsize value corr.\ with seeds 0.80--0.84}} \\
\quad {\scriptsize(r1, in 4/4 seed top-10)} But the power it would need to guarantee that no one is poor would be so great it could crush the natural rights and liberty of individuals. \\
\quad {\scriptsize(r5, in 4/4 seed top-10)} The revisionists, such as former Minister of Education Fujio and former Minister of Justice Nagano, are so zealous that they keep whitewashing Japan's role in the war even at the risk of losing office (in fact, their insistence that little killing took place in Nanking is correct: the \textbf{worst} massacres happened in the suburbs of Nanking). \\
\quad {\scriptsize(r10, in 3/4 seed top-10)} \ldots e simple: all wars betray noble purposes, militarism is the\textbf{ worst} enemy of democracy, and we mus \ldots\textsuperscript{\textdagger} \\
\bottomrule
\end{tabular}
\caption{Latent 7393: examples at activation ranks 1/5/10 per model under its own solo
vote; the activating word is bold; ($r$, in $k$/4 top-10) gives the example's rank and
how many models share the position. Top-50 activations span 30--32 distinct passages per model.}
\label{tab:huntex7393}
\end{table*}

\begin{table*}[p]
\centering
\footnotesize
\begin{tabular}{p{0.96\textwidth}}
\toprule
\textbf{\#27599} - epistemic hedging: conjecture and possibility \\
\midrule
\multicolumn{1}{l}{\emph{Qwen3}} \\
\quad {\scriptsize(r1, in 4/4 top-10)} All three were calibrated using the same unit of measurement, the cubit; the Egyptians broke the cubit into smaller units, which allowed them to keep remarkably accurate records, \textbf{perhaps} more accurate than would have been warranted for the purposes of merely agriculture and taxation. \\
\quad {\scriptsize(r5, in 4/4 top-10)} But, Namibian elephants are of great interest to tourists, and this may \textbf{be} the key to their salvation in this country that has been described as the land that God created in anger. \\
\quad {\scriptsize(r10, in 2/4 top-10)} But, Namibian elephants are of great interest to tourists, and this may be the key to their salvation in this country that has been described as the land that God created in anger. \\
\multicolumn{1}{l}{\emph{Llama-3.2}} \\
\quad {\scriptsize(r1, in 4/4 top-10)} The American Blacks played a crucial role in the very beginning of the Sydney steel plant, but never stayed around to reap any benefits. \\
\quad {\scriptsize(r5, in 2/4 top-10)} Seventy percent of the cases have occurred in eight states: Texas, Mississippi, South Dakota, Michigan, California, Louisiana, Oklahoma and Illinois. \\
\quad {\scriptsize(r10, in 1/4 top-10)} \ldots hat determines how much coverage a climate study gets? It\textbf{ probably} goes without saying that it i \ldots\textsuperscript{\textdagger} \\
\multicolumn{1}{l}{\emph{OLMo-2}} \\
\quad {\scriptsize(r1, in 3/4 top-10)} \ldots ivestock interests in the US to market in grazing permits may\textbf{ be} ive here). But putting that to\textsuperscript{\textdagger} \\
\quad {\scriptsize(r5, in 2/4 top-10)} Reinecke (U.S. Fish and Wildlife Service, Patuxent Wildlife Research Center) that contains over 5,000 citations. \\
\quad {\scriptsize(r10, in 1/4 top-10)} Physicians may now \textbf{be} more assertive about watchful waiting and follow-ups when a child's ear infection isn't severe. That may not comfort the parent of a crying child in pain, but it may be the best approach for the child in the long run. \\
\multicolumn{1}{l}{\emph{Gemma-3}} \\
\quad {\scriptsize(r1, in 2/4 top-10)} Seventy percent of the cases have occurred in eight states: Texas, Mississippi, South Dakota, Michigan, California, Louisiana, Oklahoma and Illinois. \\
\quad {\scriptsize(r5, in 3/4 top-10)} Yet, public health indicators will continue to decline unless the built environment is changed and cities are already financially constrained. Dr. Fleming offers some hope, arguing that the Affordable Care Act may \textbf{have} unlocked a key to force the private sector to recalibrate their own financial calculus and compel them to directly intervene in reducing their patients' exposure to unwalkable streets, unsafe speeds an \\
\quad {\scriptsize(r10, in 2/4 top-10)} But, Namibian elephants are of great interest to tourists, and this may be the key to their salvation in this country that has been described as the land that God created in anger. \\
\multicolumn{1}{l}{\emph{Qwen3.5-2B (adapted)}\hfill {\scriptsize value corr.\ with seeds 0.75--0.82}} \\
\quad But, Namibian elephants are of great interest to tourists, and this may be the key to their salvation in this country that has been described as the land that God created in anger. \\
\quad The American Blacks played a crucial role in the very beginning of the Sydney steel plant, but never stayed around to reap any benefits. \\
\quad Fish and Wildlife Service, Patuxent Wildlife Research Center) that contains over 5,000 citations. \\
\multicolumn{1}{l}{\emph{Llama-3.2-1B-Instruct (adapted)}\hfill {\scriptsize value corr.\ with seeds 0.74--0.86}} \\
\quad The American Blacks played a crucial role in the very beginning of the Sydney steel plant, but never stayed around to reap any benefits. \\
\quad But, Namibian elephants are of great interest to tourists, and this may \textbf{be} the key to their salvation in this country that has been described as the land that God created in anger. \\
\quad Whenever \textbf{I} sense a face emerging from the randomness of the world, I like to sketch it, accentuating the pareidolia just slightly. Maybe I'm going crazy, but last week I saw a face in the dormer windows of a building, and did this quick sketch to push it just a little. \\
\multicolumn{1}{l}{\emph{Hunyuan-1.8B (adapted, cross-lab)}\hfill {\scriptsize value corr.\ with seeds 0.76--0.79}} \\
\quad {\scriptsize(r1, in 4/4 seed top-10)} But, Namibian elephants are of great interest to tourists, and this may \textbf{be} the key to their salvation in this country that has been described as the land that God created in anger. \\
\quad {\scriptsize(r5, in 1/4 seed top-10)} In another paper Pearson acknowledges that the absence of punitive marginal tax rates is \textbf{probably} not an important consideration when people in Cape York Peninsula make their decisions about how many hours of the week they allocate to work or leisure. \\
\quad {\scriptsize(r10, in 4/4 seed top-10)} But, Namibian elephants are of great interest to tourists, and this may be the key to their salvation in this country that has been described as the land that God created in anger. \\
\bottomrule
\end{tabular}
\caption{Latent 27599: examples at activation ranks 1/5/10 per model under its own solo
vote; the activating word is bold; ($r$, in $k$/4 top-10) gives the example's rank and
how many models share the position. Top-50 activations span 35--40 distinct passages per model.}
\label{tab:huntex27599}
\end{table*}

\begin{table*}[p]
\centering
\footnotesize
\begin{tabular}{p{0.96\textwidth}}
\toprule
\textbf{\#23065} - conditional mood in hypothetical clauses \\
\midrule
\multicolumn{1}{l}{\emph{Qwen3}} \\
\quad {\scriptsize(r1, in 2/4 top-10)} Simulations are supporting the education system to better equip students with practical knowledge, so when it comes to tackling real-life situations in their future careers they can perform better. Taking into consideration the benefits offered by simulations in the education process, it \textbf{would} be a shame to miss out on them. \\
\quad {\scriptsize(r5, in 3/4 top-10)} The feature was named by Blanchard and Davis,66 who called it the Bone Springs arch. It \textbf{would} seem from their paper that they considered the feature to be anticlinal, and to have a similar, opposing flank to the north. \\
\quad {\scriptsize(r10, in 3/4 top-10)} Thoth - Link With Sirius Sirius Mythology - Zep Tepi Star - Star Tetrahedron Thoth The Scribe used sacred geometry - geometry of the stars to write our program placing coded messages in the heavens for you to find. Creators - Guardians - 'God/Dog' - Dog Star - \textbf{would} live beyond the stars in a higher realm of consciousness. \\
\multicolumn{1}{l}{\emph{Llama-3.2}} \\
\quad {\scriptsize(r1, in 3/4 top-10)} Didn't scientific research recently concluded that when in the dark (can't be seen!), people are more likely to steal? It \textbf{would} seem that we all agree on the moral high grounds but consistently fail when faced such tests. \\
\quad {\scriptsize(r5, in 3/4 top-10)} The feature was named by Blanchard and Davis,66 who called it the Bone Springs arch. It \textbf{would} seem from their paper that they considered the feature to be anticlinal, and to have a similar, opposing flank to the north. \\
\quad {\scriptsize(r10, in 2/4 top-10)} The army was given prime importance however the nation knew that it was also important for the country to protect itself from the sky and the water which lead to buying a beast to guard our oceans. \\
\multicolumn{1}{l}{\emph{OLMo-2}} \\
\quad {\scriptsize(r1, in 3/4 top-10)} The feature was named by Blanchard and Davis,66 who called it the Bone Springs arch. It \textbf{would} seem from their paper that they considered the feature to be anticlinal, and to have a similar, opposing flank to the north. \\
\quad {\scriptsize(r5, in 3/4 top-10)} Thoth - Link With Sirius Sirius Mythology - Zep Tepi Star - Star Tetrahedron Thoth The Scribe used sacred geometry - geometry of the stars to write our program placing coded messages in the heavens for you to find. Creators - Guardians - 'God/Dog' - Dog Star - \textbf{would} live beyond the stars in a higher realm of consciousness. \\
\quad {\scriptsize(r10, in 3/4 top-10)} If we will heal and open our eyes through meditation we shall indeed see God. \\
\multicolumn{1}{l}{\emph{Gemma-3}} \\
\quad {\scriptsize(r1, in 1/4 top-10)} \ldots  in vain forlet to give her signs of affection, and Horatio\textbf{ would} have little reason to think th \ldots\textsuperscript{\textdagger} \\
\quad {\scriptsize(r5, in 3/4 top-10)} If we will heal and open our eyes through meditation we shall indeed see God. \\
\quad {\scriptsize(r10, in 1/4 top-10)} And 2,000 was a terrific accomplishment. CONAN: And the number of errors he recorded even as a great defensive catcher \textbf{would} have been, you know, totally unacceptable by today's standards. \\
\multicolumn{1}{l}{\emph{Qwen3.5-2B (adapted)}\hfill {\scriptsize value corr.\ with seeds 0.95--0.98}} \\
\quad If you have any questions, don't hesitate to contact me at firstname.lastname@example.org My Kids Turn is a website which hosts 6 different shows which provides quick ideas for parents to use with their children. \\
\quad The feature was named by Blanchard and Davis,66 who called it the Bone Springs arch. It \textbf{would} seem from their paper that they considered the feature to be anticlinal, and to have a similar, opposing flank to the north. \\
\quad The industry at this period (not much better today) is 6:1, i.e.; the thickness of the board must be no more than 6 times the diameter of the hole. This ratio \textbf{would} dominate the size of the board. \\
\multicolumn{1}{l}{\emph{Llama-3.2-1B-Instruct (adapted)}\hfill {\scriptsize value corr.\ with seeds 0.95--0.99}} \\
\quad The feature was named by Blanchard and Davis,66 who called it the Bone Springs arch. It \textbf{would} seem from their paper that they considered the feature to be anticlinal, and to have a similar, opposing flank to the north. \\
\quad If you have any questions, don't hesitate to contact me at firstname.lastname@example.org My Kids Turn is a website which hosts 6 different shows which provides quick ideas for parents to use with their children. \\
\quad Yorick, the dead jester whose skull Hamlet holds during this scene, is said to have been in the earth "three-and-twenty years," which \textbf{would} make Hamlet no more than seven years old when he last rode on Yorick's back. \\
\multicolumn{1}{l}{\emph{Hunyuan-1.8B (adapted, cross-lab)}\hfill {\scriptsize value corr.\ with seeds 0.96--0.98}} \\
\quad {\scriptsize(r1, in 4/4 seed top-10)} The poor level of their response is not surprising, but it does exemplify the tactics of the whole `bury ones head in the sand'' movement - they'd much rather make noise than actually work out what is happening. It \textbf{would} be nice if this demonstration of intellectual bankruptcy got some media attention itself. \\
\quad {\scriptsize(r5, in 4/4 seed top-10)} \ldots h make for a controversial issue. From my point of view, it\textbf{ would} be ideal if one day we no\textsuperscript{\textdagger} \\
\quad {\scriptsize(r10, in 1/4 seed top-10)} This one's "skin" pattern is camouflage, but it looks a lot like some venomous western rattlers I've seen. Even without fear of snakes, this \textbf{would} still give one pause if you were trapped and couldn't move. \\
\bottomrule
\end{tabular}
\caption{Latent 23065: examples at activation ranks 1/5/10 per model under its own solo
vote; the activating word is bold; ($r$, in $k$/4 top-10) gives the example's rank and
how many models share the position. Top-50 activations span 40--43 distinct passages per model.}
\label{tab:huntex23065}
\end{table*}

\begin{table*}[p]
\centering
\footnotesize
\begin{tabular}{p{0.96\textwidth}}
\toprule
\textbf{\#20341} - programming operations, in code and in prose about code \\
\midrule
\multicolumn{1}{l}{\emph{Qwen3}} \\
\quad {\scriptsize(r1, in 4/4 top-10)} \ldots ned a new file via my IDE and turned it into a list via read\textbf{lines} . Why? Because I want to\textsuperscript{\textdagger} \\
\quad {\scriptsize(r5, in 4/4 top-10)} \ldots e opened a new file via my IDE and turned it into a list via\textbf{ read} lines. Why? Because I want\textsuperscript{\textdagger} \\
\quad {\scriptsize(r10, in 3/4 top-10)} When I use ['Some name\_\_\_\_\_\_\_\_\_\_1.5 6.5 6.5\textbackslash{}n', 'Another name\_\_\_\_\_\_\_\_6.3 1.2 1.5\textbackslash{}n'] 1.5 3.5 4.5 2.5 3.5 4.5 5.5 3.5 4.5 fileFolder = open('TEXTFILE', 'r') readFile = \textbf{fileFolder.readlines}() for line in readFile: line = line.split("\_") grades = line[-1] grades = map(float,line[-1]) could not convert string to float: . \\
\multicolumn{1}{l}{\emph{Llama-3.2}} \\
\quad {\scriptsize(r1, in 4/4 top-10)} map is operating on each character of the string, because when you iterate over a string you get individual characters. \\
\quad {\scriptsize(r5, in 4/4 top-10)} \ldots s, separated byicolons. x :=5 is a statement. MyObject.Do\textbf{Something} is a statement. Even aend\textsuperscript{\textdagger} \\
\quad {\scriptsize(r10, in 3/4 top-10)} When I use ['Some name\_\_\_\_\_\_\_\_\_\_1.5 6.5 6.5\textbackslash{}n', 'Another name\_\_\_\_\_\_\_\_6.3 1.2 1.5\textbackslash{}n'] 1.5 3.5 4.5 2.5 3.5 4.5 5.5 3.5 4.5 fileFolder = open('TEXTFILE', 'r') readFile = \textbf{fileFolder.readlines}() for line in readFile: line = line.split("\_") grades = line[-1] grades = map(float,line[-1]) could not convert string to float: . \\
\multicolumn{1}{l}{\emph{OLMo-2}} \\
\quad {\scriptsize(r1, in 4/4 top-10)} map is operating on each character of the string, because when you iterate over a string you get individual characters. \\
\quad {\scriptsize(r5, in 4/4 top-10)} \ldots e opened a new file via my IDE and turned it into a list via\textbf{ read} lines. Why? Because I want\textsuperscript{\textdagger} \\
\quad {\scriptsize(r10, in 3/4 top-10)} When I use ['Some name\_\_\_\_\_\_\_\_\_\_1.5 6.5 6.5\textbackslash{}n', 'Another name\_\_\_\_\_\_\_\_6.3 1.2 1.5\textbackslash{}n'] 1.5 3.5 4.5 2.5 3.5 4.5 5.5 3.5 4.5 fileFolder = open('TEXTFILE', 'r') readFile = \textbf{fileFolder.readlines}() for line in readFile: line = line.split("\_") grades = line[-1] grades = map(float,line[-1]) could not convert string to float: . \\
\multicolumn{1}{l}{\emph{Gemma-3}} \\
\quad {\scriptsize(r1, in 4/4 top-10)} \ldots ned a new file via my IDE and turned it into a list via read\textbf{lines} . Why? Because I want to\textsuperscript{\textdagger} \\
\quad {\scriptsize(r5, in 2/4 top-10)} For example, functions (ie, procedures that return a value) could not be called without making use of that return value. \\
\quad {\scriptsize(r10, in 1/4 top-10)} exclusive NOR), respectively. These macros \textbf{expand} into calls of binary functions such as binary-logand, binary-logior, etc. \\
\multicolumn{1}{l}{\emph{Qwen3.5-2B (adapted)}\hfill {\scriptsize value corr.\ with seeds 0.64--0.76}} \\
\quad I like to do instead a CTRL+X which is a \textbf{CUT} and then click where I want to move it to and hit CTRL+V to paste. \\
\quad I like to do instead a CTRL+X which is a \textbf{CUT} and then click where I want to move it to and hit CTRL+V to paste. \\
\quad \ldots e I want to move it to and hit CTRL+V to paste. This CUT and\textbf{ASTE} will always work and that\textsuperscript{\textdagger} \\
\multicolumn{1}{l}{\emph{Llama-3.2-1B-Instruct (adapted)}\hfill {\scriptsize value corr.\ with seeds 0.68--0.80}} \\
\quad I like to do instead a CTRL+X which is a \textbf{CUT} and then click where I want to move it to and hit CTRL+V to paste. \\
\quad I like to do instead a CTRL+X which is a \textbf{CUT} and then click where I want to move it to and hit CTRL+V to paste. \\
\quad \ldots e I want to move it to and hit CTRL+V to paste. This CUT and\textbf{ASTE} will always work and that\textsuperscript{\textdagger} \\
\multicolumn{1}{l}{\emph{Hunyuan-1.8B (adapted, cross-lab)}\hfill {\scriptsize value corr.\ with seeds 0.67--0.77}} \\
\quad {\scriptsize(r1, in 4/4 seed top-10)} \ldots ack which is CTRL+Z. I like to do instead a CTRLX which is a\textbf{ CUT} and then click where I want to\textsuperscript{\textdagger} \\
\quad {\scriptsize(r5, in 2/4 seed top-10)} In Win32 you can use different conventions (ways of passing parameters): \textbf{\_\_stdcall}, \_\_cdecl, \_\_fastcall etc. \\
\quad {\scriptsize(r10, in 3/4 seed top-10)} \ldots verticalLS. I select all these, I right click on it, I hit\textbf{ INSERT} and I shift them to the right\textsuperscript{\textdagger} \\
\bottomrule
\end{tabular}
\caption{Latent 20341: examples at activation ranks 1/5/10 per model under its own solo
vote; the activating word is bold; ($r$, in $k$/4 top-10) gives the example's rank and
how many models share the position. Top-50 activations span 8--9 distinct passages per model; this latent concentrates in the window's few code-bearing documents (largest passage up to 44\% of its top 50).}
\label{tab:huntex20341}
\end{table*}

\begin{table*}[p]
\centering
\footnotesize
\begin{tabular}{p{0.96\textwidth}}
\toprule
\textbf{\#30007} - sentence-initial imperatives: calls to action \\
\midrule
\multicolumn{1}{l}{\emph{Qwen3}} \\
\quad {\scriptsize(r1, in 4/4 top-10)} Glyptodonts became extinct at the end of the last ice age along with a large number of other megafaunal species, including pampatheres, the giant ground sloths, and the Macrauchenia. Finding an ancient shell like this is surely not an ordinary discovery. \\
\quad {\scriptsize(r5, in 1/4 top-10)} The second class is known as ``blockchain technology'', and consists of both cryptocurrencies like Bitcoin and Ethereum, and alternative distributed ledger designs like Hyperledger and ErisDB. \\
\quad {\scriptsize(r10, in 3/4 top-10)} If this ratio is used to design the board the board size would be increased in area by greater than 9 times. \\
\multicolumn{1}{l}{\emph{Llama-3.2}} \\
\quad {\scriptsize(r1, in 2/4 top-10)} \ldots d her bed, pulling down the covers, and patting the pillow).\textbf{Give} your child some advance notice  \ldots\textsuperscript{\textdagger} \\
\quad {\scriptsize(r5, in 3/4 top-10)} \ldots o list here and too many to keep up with as they come and go.\textbf{Use} this Google web search form to  \ldots\textsuperscript{\textdagger} \\
\quad {\scriptsize(r10, in 1/4 top-10)} Keep us clear from ungodly activities Lord that we may not expose ourselves to unneeded trouble but improve our quality of life and walk in You. The opinions expressed by authors may not necessarily reflect the opinion of FaithWriters.com. \\
\multicolumn{1}{l}{\emph{OLMo-2}} \\
\quad {\scriptsize(r1, in 1/4 top-10)} If students are given the chance to try out what they have learned theoretically, they will have a practical overview on the material and therefore they will get a deeper understanding on how things work in reality. \\
\quad {\scriptsize(r5, in 4/4 top-10)} Glyptodonts became extinct at the end of the last ice age along with a large number of other megafaunal species, including pampatheres, the giant ground sloths, and the Macrauchenia. Finding an ancient shell like this is surely not an ordinary discovery. \\
\quad {\scriptsize(r10, in 1/4 top-10)} - respiratory failure - liver failure - heart failure - myelodysplastic syndromes \textbf{Call} Health Care Provider: Call your health care provider if you suspect you or someone you have had close contact with has SARS. \\
\multicolumn{1}{l}{\emph{Gemma-3}} \\
\quad {\scriptsize(r1, in 1/4 top-10)} \ldots t the age of6 after atracted battle with his failing health.\textbf{Write} the first section of your page \ldots\textsuperscript{\textdagger} \\
\quad {\scriptsize(r5, in 1/4 top-10)} And so that really fed into the whole icon of them, the whole idea of them as American folk heroes. CONAN: Folk heroes. \\
\quad {\scriptsize(r10, in 1/4 top-10)} \ldots ks toage. More eyes and ears can help to thwart predators.\textbf{ Listen} to a noisy, mixreed flock\textsuperscript{\textdagger} \\
\multicolumn{1}{l}{\emph{Qwen3.5-2B (adapted)}\hfill {\scriptsize value corr.\ with seeds 0.85--0.87}} \\
\quad If this ratio is used to design the board the board size would be increased in area by greater than 9 times. \\
\quad \ldots scovery tools, such as Summon (which Penrose Library uses).\textbf{ Think} of it this way:\textsuperscript{\textdagger} \\
\quad \ldots dent of red blood cell production, this may be possible too.\textbf{ Make} a drug that stimulates the\textsuperscript{\textdagger} \\
\multicolumn{1}{l}{\emph{Llama-3.2-1B-Instruct (adapted)}\hfill {\scriptsize value corr.\ with seeds 0.84--0.91}} \\
\quad \ldots dent of red blood cell production, this may be possible too.\textbf{ Make} a drug that stimulates the\textsuperscript{\textdagger} \\
\quad \ldots d her bed, pulling down the covers, and patting the pillow).\textbf{Give} your child some advance notice\textsuperscript{\textdagger} \\
\quad Currently the company is facing financial woes and will hopefully once again make back to the top of the auto industry. \\
\multicolumn{1}{l}{\emph{Hunyuan-1.8B (adapted, cross-lab)}\hfill {\scriptsize value corr.\ with seeds 0.85--0.86}} \\
\quad {\scriptsize(r1, in 4/4 seed top-10)} \ldots dent of red blood cell production, this may be possible too.\textbf{ Make} a drug that stimulates the bra \ldots\textsuperscript{\textdagger} \\
\quad {\scriptsize(r5, in 0/4 seed top-10)} By allowing your toddler to make limited choices, she'll feel empowered (and you'll be satisfied with the result). \\
\quad {\scriptsize(r10, in 0/4 seed top-10)} While Jesus' words of warning in the Gospel might seem stern, they are not nearly as harsh as the same message from Mark (\textbf{read} last Advent). \\
\bottomrule
\end{tabular}
\caption{Latent 30007: examples at activation ranks 1/5/10 per model under its own solo
vote; the activating word is bold; ($r$, in $k$/4 top-10) gives the example's rank and
how many models share the position. Top-50 activations span 38--40 distinct passages per model.}
\label{tab:huntex30007}
\end{table*}